\documentclass[11pt]{article}
\usepackage{amsmath, amssymb}
\usepackage{graphicx}
\usepackage{caption}
\usepackage{subcaption}
\usepackage{geometry}
\usepackage{authblk}
\usepackage{cite}
\usepackage{float}
\usepackage{hyperref}
\usepackage{multirow}
\usepackage{booktabs,tabularx}
\title{Comparative Study of Weighted and Coupled Second- and Fourth-Order PDEs
for Image Despeckling in Grayscale, Color, SAR, and Ultrasound}

  \author{
    \textbf{Manish Kumar$^{1*}$,Rajendra K. Ray $^{2}$} \\
    \textsuperscript{1,2} School of Mathematical and Statistical Sciences, \\
    Indian Institute of Technology Mandi, Mandi (H.P), 175075, India \\
    \texttt{manishkumarkitlana@gmail.com}, \texttt{rajendra@iitmandi.ac.in} \\
    \textsuperscript{*}\footnote{Corresponding author: Rajendra K. Ray, \texttt{rajendra@iitmandi.ac.in}}}
\date{}

\begin{document}

\maketitle

\begin{abstract}
Partial Differential Equation (PDE)-based approaches have gained significant attention in image despeckling due to their strong capability to preserve structural details while suppressing noise. However, conventional second-order PDE models tend to generate blocky artifacts, whereas higher-order models often introduce speckle patterns. To resolve it, this paper proposes and comparatively analyzes two advanced PDE-based frameworks designed for speckle noise suppression while preserving the fine edges. The first model introduces a novel weighted formulation that combines second and fourth-order PDEs through a weighting parameter. The second-order diffusion coefficient employs grayscale and gradient-based indicators, while the fourth-order term is guided solely by a Laplacian-based indicator. The second model constructs a coupled PDE framework, where independent fourth and second-order components are explicitly solved in an iterative manner. In this coupled structure, each diffusion coefficient is defined separately to enhance adaptability in varying image regions.
Both models are implemented using the explicit finite difference method. The proposed techniques are extensively evaluated on a variety of datasets, including standard grayscale, color, Synthetic Aperture Radar (SAR), and ultrasound images. Comparative experiments with the existing Telegraph Diffusion Model (TDM) and Fourth-Order Telegraph Diffusion Model (TDFM) demonstrate the superiority of the proposed approaches in reducing speckle noise while effectively preserving fine image structures and edges. Quantitative evaluations using PSNR, SSIM and Speckle Index metrics confirm that the proposed models produce higher image quality and enhanced visual perception. Overall, the presented PDE-based formulations provide a reliable and efficient framework for image despeckling in both natural and medical imaging.
\end{abstract}

\section{Introduction}
In imaging applications, acquired images are frequently degraded by various forms of noise, which is often an inevitable consequence of sensor imperfections, environmental disturbances, and acquisition modality. Among the challenging types of noise, speckle noise is especially prominent in coherent imaging systems such as synthetic aperture radar (SAR) and ultrasound, severely affecting the ability to interpret and analyze. The generation of speckle noise is typically modeled as a multiplicative process.
\begin{equation}
    I = I_0 \cdot \eta,
\end{equation}
where $I_0$ denotes the underlying true image and $\eta$ is a multiplicative random variable that represents the speckle noise~\cite{1}. In real SAR and ultrasound scenarios, this multiplicative noise is well described using a Gamma distribution characterized by the number of looks $L$, which determines the balance between spatial resolution and noise variance~\cite{2,5,6}. The fundamental challenge faced by despeckling methods is to robustly suppress speckle while preserving critical features such as edges, textures, and structural details.
To address this challenge, researchers have developed a variety of despeckling models, each based on unique mathematical and algorithmic foundations. Traditionally, spatial domain filtering lays the foundation for noise suppression, employing operations such as mean, median, and adaptive filters to locally average pixel intensities. Although these methods can noticeably reduce noise, they frequently compromise resolution, resulting in blurred edges and loss of fine structures. Multi-look averaging is a fundamental technique that reduces noise while sacrificing spatial detail~\cite{47}, while adaptive schemes improve on this by considering local variance or edge information~\cite{8}. However, comprehensive spatial methodologies are subject to excessive smoothing, especially in highly textured or edge-rich regions.
Wavelet and transform-based methodologies represent a further category, using multi-scale decompositions to differentiate noise from significant image content. Techniques like wavelet shrinkage and wavelet-based MAP estimators can efficiently reduce noise and recover concealed features~\cite{21,25,36}. However, these models are subject to artifacts like ringing, and their effectiveness depends on the selection of the basis and thresholding criteria. Homomorphic filtering—achieved through logarithmic transformation of the image, denoising in the log-domain, and subsequently exponentiation—allows the transformation of the multiplicative speckle model into an additive one. This method has limitations, as the non-linearity of the logarithmic transformation may create bias and increase intensity variations~\cite{2}.
Patch-based and non-local denoising algorithms have substantially advanced the exploitation of self-similarity properties inherent in natural images. Approaches such as Non-Local Means (NLM)~\cite{16,17,18}, PPB~\cite{18}, and BM3D~\cite{50} perform weighted averaging of similar image patches across the entire domain, rather than restricting computations to local neighborhoods. This global similarity-driven strategy enables improved restoration of fine textures and edges compared to traditional local smoothing filters. In the context of speckle noise reduction, specialized variants like SAR-BM3D~\cite{16} have been developed to accommodate the statistical characteristics of multiplicative noise. Despite their impressive performance, these methods face challenges related to high computational complexity and potential over-smoothing of low-contrast or weakly textured regions, where establishing reliable patch correspondence is difficult.
A parallel line of progress in image despeckling has emerged through variational and PDE-based frameworks, which convert the denoising task into an optimization problem that balances data fidelity against appropriate regularization constraints~\cite{34,35,36,37,38,39,40}. Among these, total variation (TV) regularization~\cite{39} remains one of the most influential techniques, known for its ability to preserve sharp edges while effectively reducing noise. However, the tendency of TV models to produce piecewise-constant solutions often results in undesirable staircase artifacts, and their performance degrades in regions with rich textures or fine structural details.  
Also, nonlinear PDE models have been introduced as flexible and powerful tools ~\cite{31,32,34}. Second-order PDE formulations typically yield block-like artifacts, whereas higher-order schemes, particularly those incorporating fourth-order derivatives~\cite{29}, enhance smoothness by maintaining gradual transitions and suppressing piecewise-plane effects. However, these advanced formulations often introduce speckle effects.
The deep learning models have also improved image denoising, with convolutional neural networks (CNNs) and generative models achieving state-of-the-art results on benchmark datasets~\cite{41,42,43,44,45,46,65}. Despite their success, these models require extensive labeled data, high computational resources, and may struggle to generalize to unseen noise distributions. Their limited interpretability poses additional challenges in sensitive domains such as medical or SAR imaging.
In SAR imaging, where true reference images are rarely available, conventional quality metrics like PSNR and SSIM are inadequate. Alternative indices, such as the Speckle Index, are therefore employed for objective evaluation~\cite{5,6}.
Overall, despite the diversity of existing despeckling strategies, each approach faces inherent trade-offs among restoration quality, texture preservation, computational complexity, and robustness. Consequently, diffusion-based methods and the design of adaptive diffusion coefficients continue to attract significant research attention for achieving an improved balance between effective denoising and structural fidelity.

In 1990, Perona and Malik~\cite{70} introduced first anisotropic diffusion for the processing of images corrupted by additive Gaussian noise. The governing equation is given by 

\begin{equation}
    \frac{\partial I}{\partial t} = \text{div}\big( C(|\nabla I|) \nabla I \big),
\end{equation}

where $I$ denotes the image intensity, $t$ is the diffusion time, and $C(|\nabla I|)$ is a diffusion controlling function dependent on the magnitude of the image gradient. This function is monotonically decreasing, with properties $C(0) = 1$ and $C(x) \to 0$ as $x \to \infty$. While this model helps retain significant edges, it has limitations such as edge information loss and the potential amplification of high-frequency noise. 
Recently, Zhou et al.~\cite{72} developed the Doubly Degenerate (DD) diffusion framework, based on grayscale intensities and their spatial gradients. The DD model is governed by

\begin{equation}
    \frac{\partial I}{\partial t} = \text{div}\big( C(I, |\nabla I|) \nabla I \big), \qquad (x, t) \in \Omega \times (0, T),
\end{equation}

subject to Neumann-type (zero-flux) and initial conditions:

\begin{align}
    \frac{\partial I}{\partial n} &= 0, \hspace{1em} (x, t) \in \partial \Omega \times (0, T),\\
    I(x, 0) &= f(x), \hspace{1em} x \in \Omega,
\end{align}

where $f(x)$ is the observed noisy input and $n$ is the outward unit normal on the domain boundary. The diffusion coefficient is defined as

\begin{equation}
    C(I, |\nabla I|) = \frac{2|I|^{\alpha}}{M^{\alpha} + |I|^{\alpha}} \cdot \frac{1}{\big(1 + |\nabla I|^2\big)^{\frac{1-\beta}{2}}},
\end{equation}

with parameters $\alpha, \beta > 0$ and $M$ being a maximum of intensity of I. Here, $\beta$ controls the degree of sensitivity to gradients. This model better balances noise suppression and preservation of edges.
For further improvement, Zhou et al.~\cite{71} presented the ZZDB model, incorporating a spatially and intensity-adaptive exponent function. The corresponding PDE is

\begin{equation}
    \frac{\partial I}{\partial t} = \text{div} \left( \frac{\nabla I}{1 + \left( \frac{|\nabla I_\xi|}{K} \right)^{\beta(I)}} \right), \qquad (x, t) \in \Omega \times (0, T),
\end{equation}

where the exponent $\beta(I)$ is given by

\begin{equation}
    \beta(I) = 2 - \frac{2|I|^{\alpha}}{M^{\alpha} + |I|^{\alpha}},
\end{equation}

and $I_\xi$ denotes a smoothed version of $I$ obtained by Gaussian convolution. The same Neumann and initial conditions as above are applied. By allowing gradient sensitivity to vary with local intensity, this method achieves denoising and detail preservation.

To address the well-posedness and stability of DD-based models, Shan et al.~\cite{73} introduced a regularized approach using Gaussian-smoothed images:

\begin{equation}
    \frac{\partial I}{\partial t} = \nabla \cdot \big( C(I_\xi, |\nabla I_\xi|) \nabla I \big),
\end{equation}

with boundary and initial constraints as before. Here, $I_\xi$ is a smoothed version of I, and the diffusion coefficient

\begin{equation}
    C(I_\xi, |\nabla I_\xi|) = \left( \frac{I_\xi}{M_\xi} \right)^{\alpha} \cdot \frac{1}{1 + |\nabla I_\xi|^{\nu}},
\end{equation}

offers both intensity and gradient adaptation.

Later, Majee et al.~\cite{29} introduced a diffusion-wave (TDE) framework. The governing equation is given by 
\begin{equation}
    \frac{\partial^2 I}{\partial t^2} + \gamma \frac{\partial I}{\partial t} = \nabla \cdot \big( C(I_\xi, |\nabla I_\xi|) \nabla I \big),
\end{equation}

featuring a damping parameter $\gamma$. With the initial conditions  $I(x, 0) = f(x)$ and the time derivative $\frac{\partial I}{\partial t}(x, 0) = 0$. The diffusion coefficient is

\begin{equation}
    C(I_\xi, |\nabla I_\xi|) = \frac{2|I_\xi|^{\alpha}}{M_\xi^{\alpha} + |I_\xi|^{\alpha}} \cdot \frac{1}{1 + (|\nabla I_\xi|/k)^2},
\end{equation}

where $M_\xi = \max_{x \in \Omega} |I_\xi(x, t)|$, $\alpha$ shapes intensity adaptation, and $k$ modulates gradient response. The hyperbolic (wave) behavior helps to handle the variation in intensity.

Jain et al.~\cite{28} proposed the Coupled Partial Differential Equation (MCPDE) model, featuring a set of three interconnected equations:
\begin{align}
    \frac{\partial I}{\partial t} &= \nabla \cdot \left( \psi(I) \, g(u) \, \nabla I \right) - 2\lambda v, && (x,t) \in \Omega \times (0,T), \\
    \frac{\partial u}{\partial t} &= \kappa_1 \left( |\nabla I_\xi|^2 - u + \frac{\kappa_2^2}{2} \Delta u \right), && (x,t) \in \Omega \times (0,T), \\
    \frac{\partial v}{\partial t} &= v - \frac{I - J}{I^2}, && (x,t) \in \Omega \times (0,T).
\end{align}
The Neumann boundary conditions,
\[
    \frac{\partial I}{\partial \mathbf{n}} = 0,\quad
    \frac{\partial u}{\partial \mathbf{n}} = 0,\quad
    \frac{\partial v}{\partial \mathbf{n}} = 0,\quad
    \text{on } \partial\Omega \times (0,T),
\]
and initial conditions are:
\[
    I(x,0) = J(x),\quad v(x,0) = 0, \quad u(x,0) = G_\xi * |\nabla J|^2, \qquad x \in \Omega.
\]
In this system, the Laplacian is denoted by $\Delta$, $g(u) = \frac{1}{1+(u/k)^2}$, $\psi(I) = \frac{2|I|^\alpha}{|I|^\alpha + M^\alpha}$, and $M = \sup_{x \in \Omega} (G_\xi * J)(x)$. The function $I_\xi$ represents a Gaussian-smoothed version of $I$. Parameter selection (including $\alpha$, $\xi$, $k$, $\kappa_1$, $\kappa_2$) and the use of auxiliary variables enable refined despeckling and improved feature preservation.
Majee~\cite{69} proposed a hyperbolic-parabolic coupled PDE system (HPCPDE) that jointly evolves image and edge indicators:
\begin{equation}
    \frac{\partial^2 I}{\partial t^2} + \gamma\frac{\partial I}{\partial t}
    - \nabla \cdot \left( \frac{1}{(1 + s^\alpha)(1 + \iota |u_\xi|^\beta)} \nabla I \right) = 0, \qquad (x,t) \in \mathcal{T},
\end{equation}
\begin{equation}
    \frac{\partial u}{\partial t} - h(|\nabla I_\xi|) + u - \frac{\nu^2}{2}\Delta u = 0, \qquad (x,t) \in \mathcal{T},
\end{equation}
with
\begin{align}
    I(x,0) &= I_0(x), \quad 
    \frac{\partial I}{\partial t}(x,0) = 0,\quad
    u(x,0) = G_\xi * |\nabla I_0(x)|^2,\quad x \in \Omega,\\
    \frac{\partial I}{\partial \mathbf{n}} &= 0,\quad 
    \frac{\partial u}{\partial \mathbf{n}} = 0, \quad \text{on } \partial\Omega \times (0,T).
\end{align}

To overcome the limitations of second-order models, You and Kaveh~\cite{68} advanced a fourth-order PDE by employing the Laplacian operator:
\begin{equation}
   \frac{\partial I}{\partial t} = -\Delta \left( C(\Delta I)\Delta I \right),
\end{equation}
solved with initial and boundary conditions
\begin{equation}
    \frac{\partial I}{\partial n} = 0, \quad \text{on } \partial \Omega_T, \qquad
    I(x, 0) = f(x), \quad \text{in } \Omega,
\end{equation}
where
\begin{equation}
    C(|\Delta I|) = \frac{1}{1+(|\Delta I|/k)^2}.
\end{equation}
Rajendra Ray\cite{74} proposed a single fourth-order telegraph PDE (TDFM) model for effective speckle noise reduction:

\begin{equation}
    \frac{\partial^2 I}{\partial t^2} + \gamma \frac{\partial I}{\partial t} 
    = -\Delta \left( C(I_\xi, |\Delta I_\xi|)\, \Delta I \right) 
    - \lambda \left( \frac{I - f}{I} \right)^2, 
    \quad (x, t) \in \Omega \times (0, T),
    \label{eq:ray_model}
\end{equation}

with the initial conditions
\begin{equation}
    I(x, 0) = f(x), \quad I_t(x, 0) = 0, \quad \forall x \in \Omega,
    \label{eq:ray_ic}
\end{equation}

and Neumann-type boundary conditions
\begin{equation}
    \frac{\partial I}{\partial n} = 0, 
    \quad \frac{\partial (\Delta I)}{\partial n} = 0, 
    \quad (x, t) \in \partial \Omega \times (0, T).
    \label{eq:ray_bc}
\end{equation}

Here, $f$ denotes the observed noisy image and $t$ represents the artificial time parameter.
The nonlinear diffusion coefficient is defined as
\begin{equation}
    C(I_\xi, |\Delta I_\xi|) 
    = \frac{2 |I_\xi|^\alpha}{M_\xi^\alpha + |I_\xi|^\alpha} 
      \cdot \frac{1}{1 + \left(\frac{|\Delta I_\xi|}{k}\right)^2},
    \tag{DC}\label{eq:ray_diffusion}
\end{equation}
where $I_\xi$ is the grayscale indicator, $M_\xi$ is the maximum value of $I_\xi$, 
$\alpha$ controls sensitivity to intensity variations, and $k$ tunes the Laplacian-based smoothing strength.

 In classical fourth-order diffusion models, speckle-like artifacts are introduced. To mitigate these effects, we enhance the smoothing by incorporating second-order diffusion. Thus, we investigate two distinct strategies: one is to form an explicitly coupled system involving both fourth- and second-order equations and tackle it with an explicit iterative solver; the second is to use a weighted sum of the fourth- and second-order terms, modulated by a tunable parameter $A$.

The performance of these weighted models depends critically on the choice of $A$, as varying this parameter alters the balance between the contributions of second- and fourth-order smoothing. Detailed parameter analysis is included in the proposed section.

\section{Proposed Method}

We propose two adaptive models for speckle noise reduction. Model 1 is a weighted hyperbolic PDE framework that combines second and fourth-order diffusion using a tuning parameter $A \in [0,1]$:

\begin{equation}
    \frac{\partial^2 I}{\partial t^2} + \gamma \frac{\partial I}{\partial t} = 
    A \,[ \nabla \cdot \left( (C_1(I_\xi)C_3(|\nabla I_\xi|) \nabla I \right)]
    - (1 - A) \, [\Delta \left( C2(|\Delta I_\xi|) \Delta I \right)]
    - \lambda \left( \frac{I - f}{I} \right)^2,
    \quad (x, t) \in \Omega \times (0, T),
    \label{proposed_1}
\end{equation}

with initial conditions

\begin{equation}
    I(x, 0) = f(x), \quad \frac{\partial I}{\partial t}(x, 0) = 0, \quad \forall x \in \Omega,
\end{equation}

and Neumann boundary condition

\begin{equation}
    \frac{\partial I}{\partial n} = 0,
     \quad \frac{\partial (\Delta I)}{\partial n} = 0, \quad \forall (x, t) \in \partial \Omega \times (0, T).
\end{equation}
Here, $I$ is the evolving image, $f$ is the given noisy image, $\gamma$ and $\lambda$ are positive parameters controlling damping and data fidelity, respectively.

We propose Model 2 as a classical hyperbolic PDE formulation, which emphasizes balanced contributions from both fourth-order and second-order diffusion effects. The mathematical formulation of Model 2 is given by the following governing equation, which explicitly integrates balanced second and fourth-order diffusion terms within a hyperbolic PDE framework.

\begin{equation}
    \frac{\partial^2 I}{\partial t^2} + \gamma \frac{\partial I}{\partial t} = 
    - \Delta \left((C_1(I_\xi)C_2(|\Delta I_\xi|)) \Delta I \right) 
    - \lambda \left( \frac{I - f}{I} \right)^2,
    \quad (x, t) \in \Omega \times (0, T),
    \label{proposed_2}
\end{equation}
and

\begin{equation}
    \frac{\partial^2 I}{\partial t^2} + \gamma \frac{\partial I}{\partial t} = 
    \nabla \cdot \left( (C_1(I_\xi)C_2(|\nabla I_\xi|) \nabla I \right),
\end{equation}
with identical initial and boundary conditions:

\begin{equation}
    I(x, 0) = f(x), \quad \frac{\partial I}{\partial t}(x, 0) = 0, \quad \forall x \in \Omega,
\end{equation}

\begin{equation}
    \frac{\partial I}{\partial n} = 0,
     \quad \frac{\partial (\Delta I)}{\partial n} = 0, \quad \forall (x, t) \in \partial \Omega \times (0, T).
\end{equation}
and the diffusion coefficient is 
\[
C_1(I_\xi) =  \frac{2 |I_\xi|^\alpha}{|I_\xi|^\alpha + M_\xi^\alpha},\quad
C_2(|\Delta I_\xi|) = \frac{1}{1 + \left(\frac{|\Delta I_\xi|}{k}\right)^2}. \quad
C_3(|\nabla I_\xi|)=\frac{1}{1 + \left(\frac{|\nabla I_\xi|}{k}\right)^2}
\]
This setup allows studying how the weighted blending of second- and fourth-order PDEs affects denoising quality, in comparison to pure forms governed separately by either second- or fourth-order diffusion processes.The diffusion coefficients $C$ are coefficient functions adaptive to local image information, typically involving Gaussian smoothing ($I_\xi$) and gradient or Laplacian magnitudes, ensuring anisotropic and spatially varying diffusion behavior.

\section{Numerical Discretizations}

We discretize the two-dimensional spatial domain $\Omega$ with grid spacings $\delta x$ (in $x$), $\delta y$ (in $y$), and a temporal step size $\delta t$. The grid points are indexed such that $I_{i,j}^n$ represents the numerical approximation of $I(x_i, y_j, t_n)$.

\[
\frac{\partial^2 I}{\partial t^2}\Big|_{i,j}^n \approx \frac{I_{i,j}^{n+1}-2I_{i,j}^n+I_{i,j}^{n-1}}{\delta t^2},\qquad
\frac{\partial I}{\partial t}\Big|_{i,j}^n \approx \frac{I_{i,j}^{n+1}-I_{i,j}^n}{\delta t}.
\]

\[
\begin{aligned}
C_{i+\tfrac12,j} &= \tfrac12(C_{i+1,j}+C_{i,j}),\qquad
C_{i,j+\tfrac12} = \tfrac12(C_{i,j+1}+C_{i,j}),\\[4pt]
\mathrm{FD}_2[C\nabla I]_{i,j} &= 
\frac{1}{\delta x^2}\Big( C_{i+\tfrac12,j}(I_{i+1,j}-I_{i,j}) - C_{i-\tfrac12,j}(I_{i,j}-I_{i-1,j})\Big)\\
&\qquad +\frac{1}{\delta y^2}\Big( C_{i,j+\tfrac12}(I_{i,j+1}-I_{i,j}) - C_{i,j-\tfrac12}(I_{i,j}-I_{i,j-1})\Big).
\end{aligned}
\]

\[
\begin{aligned}
(\Delta I)_{i,j} &= \frac{I_{i+1,j}-2I_{i,j}+I_{i-1,j}}{\delta x^2} + \frac{I_{i,j+1}-2I_{i,j}+I_{i,j-1}}{\delta y^2},\\[4pt]
B_{i,j} &= C_{i,j}\,(\Delta I)_{i,j},\\[4pt]
\mathrm{FD}_4[C\Delta I]_{i,j} &= (\Delta B)_{i,j}
= \frac{B_{i+1,j}-2B_{i,j}+B_{i-1,j}}{\delta x^2}
+ \frac{B_{i,j+1}-2B_{i,j}+B_{i,j-1}}{\delta y^2}.
\end{aligned}
\]

\[
S_{i,j}^n \;=\; \lambda\Big(\frac{I_{i,j}^n - f_{i,j}}{I_{i,j}^n + \varepsilon}\Big)^{\!2},\qquad \varepsilon>0.
\]

\[
\frac{I^{n+1}-2I^n+I^{n-1}}{\delta t^2} + \gamma\frac{I^{n+1}-I^n}{\delta t} = \mathrm{RHS}^n
\quad\Longrightarrow\quad
I_{i,j}^{n+1}
= \frac{\displaystyle \mathrm{RHS}_{i,j}^n + \dfrac{2I_{i,j}^n-I_{i,j}^{n-1}}{\delta t^2} + \dfrac{\gamma}{\delta t}\,I_{i,j}^n}
{\displaystyle \dfrac{1}{\delta t^2}+\dfrac{\gamma}{\delta t}}.
\]

\[
\mathrm{RHS}_{i,j}^n =
\begin{cases}
A\,\mathrm{FD}_2[C_1\nabla I]_{i,j}^n - (1-A)\,\mathrm{FD}_4[C_2\Delta I]_{i,j}^n - S_{i,j}^n, & \text{(Model 1)}\\[6pt]
-\,\mathrm{FD}_4[C\,\Delta I]_{i,j}^n - S_{i,j}^n, & \text{(Model 2, 4th-order)}\\[4pt]
\mathrm{FD}_2[C\nabla I]_{i,j}^n, & \text{(Model 2, 2nd-order variant).}
\end{cases}
\]

\subsection* {Extension to color image restoration}
To comprehensively evaluate the adaptability of the proposed denoising techniques, both models were implemented for the restoration of color images. The process involved decomposing the noisy RGB image into its individual red, green, and blue channels. Each channel was then independently processed with the PDE-based denoising algorithm, enabling targeted noise reduction tailored to the features of each color component. Upon completion, the three denoised channels were merged using the \texttt{cat} function to reconstruct the full color image.

For quantitative assessment, the quality of the reconstruction was measured by calculating the Peak Signal-to-Noise Ratio (PSNR) and Mean Structural Similarity Index (MSSIM) for the restored images, comparing them against the ground-truth color images. This channel-wise approach effectively retains essential image details while achieving robust noise suppression, demonstrating strong performance across a wide range of color imaging applications.

\section{Evaluation Metrics and Parameters}

To comprehensively evaluate the quality of image restoration algorithms, quantitative and perceptual criteria are used, especially when reference data is available. When ground truth images are provided—such as in synthetic experiments where noise is artificially injected—the Peak Signal-to-Noise Ratio (PSNR) serves as a fundamental metric for measuring reconstruction fidelity. PSNR is calculated using the maximum pixel intensity of the original image and the average squared error between the restored image $I^{(k)}$ and the reference $I$ across all $M \times N$ pixels:
\begin{equation}
    \text{PSNR} = 10 \log_{10} \left( \frac{\max(I)^2}
    {\frac{1}{MN} \sum_{i=1}^{M} \sum_{j=1}^{N} \left( I(i,j) - I^{(k)}(i,j) \right)^2} \right),
\end{equation}
where higher PSNR values indicate better preservation of image intensity and reduced error.

In addition to pixel-wise fidelity, structural preservation and perceptual quality are also critical. The Structural Similarity Index (SSIM) quantitatively evaluates the similarity between a restored image and its reference by comparing local patterns of luminance, contrast, and structure within moving windows. For any local window, SSIM is defined as:
\begin{equation}
    \text{SSIM}(I, I^{(k)}) = \frac{(2\mu_I \mu_{I^{(k)}} + C_1)(2\sigma_{I I^{(k)}} + C_2)}
    {(\mu_I^2 + \mu_{I^{(k)}}^2 + C_1)(\sigma_I^2 + \sigma_{I^{(k)}}^2 + C_2)},
\end{equation}
where $\mu_I$ and $\mu_{I^{(k)}}$ are local means, $\sigma_I^2$ and $\sigma_{I^{(k)}}^2$ are variances, $\sigma_{I I^{(k)}}$ denotes covariance, and $C_1$, $C_2$ are small constants that prevent division by zero. SSIM values range from 0 to 1, where values closer to 1 signify higher structural similarity with the reference.

To capture overall structural fidelity across the entire image or dataset, the mean SSIM (MSSIM) is calculated by averaging the SSIM scores over $K$ evaluation windows or iterations:
\begin{equation}
    \text{MSSIM} = \frac{1}{K} \sum_{k=1}^{K} \text{SSIM}(I, I^{(k)}).
\end{equation}
Thus, MSSIM provides a comprehensive indication of perceptual image integrity throughout the restoration process.

For real-world scenarios lacking reference images, a relative convergence criterion is used to monitor the progress of iterative restoration algorithms. The iterative process halts when the relative change between consecutive outputs drops below a pre-defined tolerance:
\begin{equation}
    \frac{\| I^{(k+1)} - I^{(k)} \|_2^2}{\| I^{(k)} \|_2^2} \leq \varepsilon,
\end{equation}
where $\varepsilon$ is set to $1 \times 10^{-4}$ in our implementations, ensuring sufficient stabilization without unnecessary computation.
\begin{figure}[H]
\centering
\begin{tabular}{cccc} 
\includegraphics[width=0.27\textwidth]{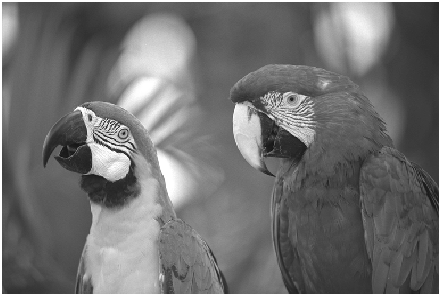} & 
\includegraphics[width=0.21\textwidth]{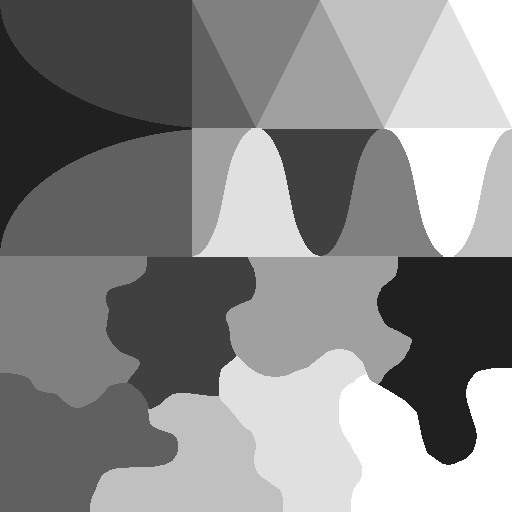} & 
\end{tabular}
\caption{Images: (a) Parrots, (b) Texture}
\label{fig:gray_images}
\end{figure}
\begin{figure}[H]
\centering
\begin{tabular}{cccc} 
\includegraphics[width=0.27\textwidth]{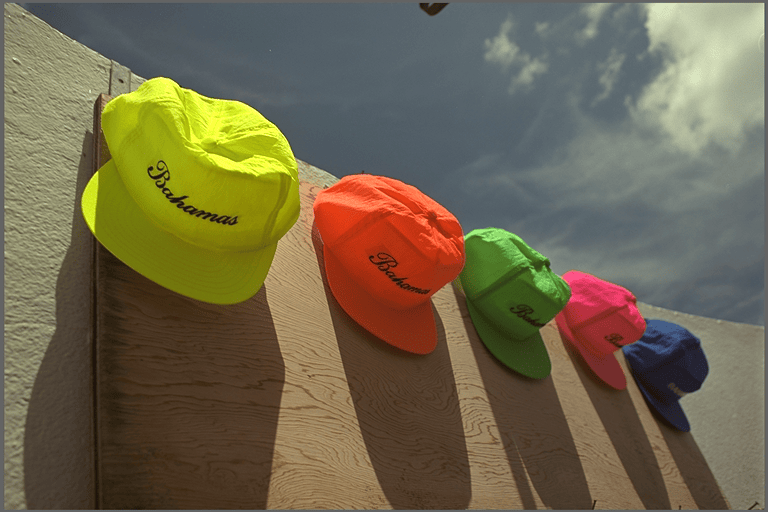} & 
\includegraphics[width=0.22\textwidth]{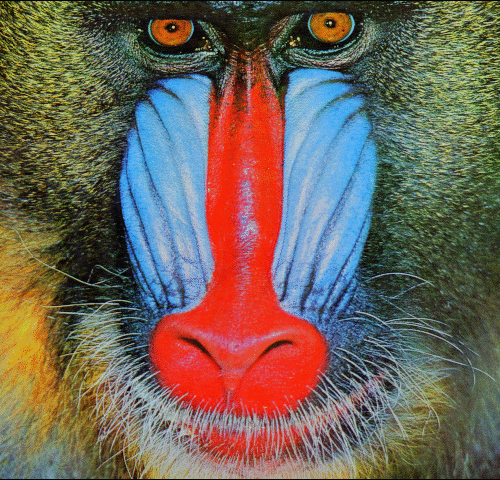}  
\end{tabular}
\caption{color Images: (a) caps, (b)baboon.}
\label{fig:color_images}
\end{figure}
The persistence and severity of speckle noise, particularly in SAR and ultrasound imaging, are evaluated using the Speckle Index (SI), a statistic reflecting image homogeneity:
\begin{equation}
    \text{SI} = \frac{\sigma}{\mu},
\end{equation}
Here, $\sigma$ represents the standard deviation and $\mu$ is the mean intensity of all the pixel values within an image. A lower SI value indicates reduced speckle noise, which corresponds to enhanced image clarity and quality, particularly significant for medical and radar imaging applications. In this study, we conducted an extensive evaluation of different image denoising models on multiple datasets varying in complexity and noise level.

To ensure consistent performance in different images, we experimented with four different values of the weighting parameter \(A\): 0.3, 0.5, 0.7, and 0.9. After careful evaluation, we observed that \(A = 0.7\) provides the best balance between edge preservation and smoothing. Furthermore, by maintaining all other parameters constant, the optimal PSNR and SSIM values are achieved at a value of 0.7. Lower values tend to under-smooth the image, while higher values may result in excessive blurring of important details. Therefore, the parameter \(A\) is set at 0.7 for all subsequent experiments.

\begin{figure}[h!]
    \centering
    \begin{subfigure}{0.19\textwidth}
        \includegraphics[width=\linewidth]{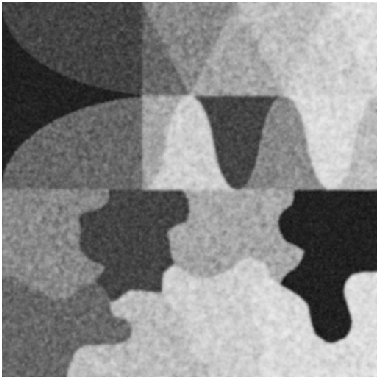}
        \caption*{\(A = 0.3\)}
    \end{subfigure}
    \begin{subfigure}{0.19\textwidth}
        \includegraphics[width=\linewidth]{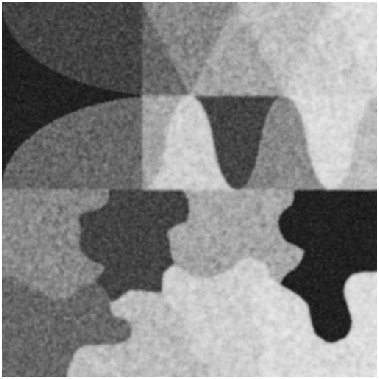}
        \caption*{\(A = 0.5\)}
    \end{subfigure}
    \begin{subfigure}{0.19\textwidth}
        \includegraphics[width=\linewidth]{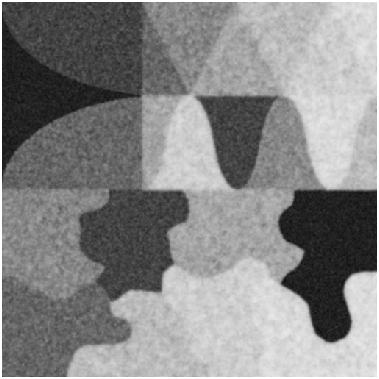}
        \caption*{\(A = 0.7\)}
    \end{subfigure}
    \begin{subfigure}{0.19\textwidth}
        \includegraphics[width=\linewidth]{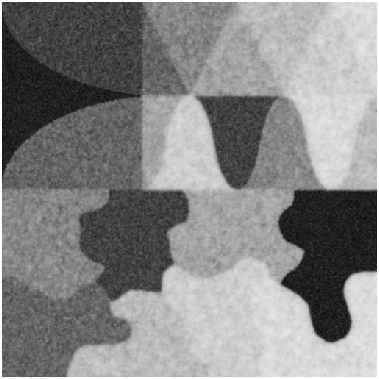}
        \caption*{\(A = 0.9\)}
    \end{subfigure}
    \caption{Effect of varying the weighting parameter \(A\) on image restoration results.}
    \label{fig:parameter_A_comparison}
\end{figure}

Figures ~\ref{fig:gray_images} and ~\ref{fig:color_images} illustrate the datasets used for this evaluation. Figure ~\ref{fig:gray_images} presents two grayscale images, identified as Parrots and Texture while Figure  ~\ref{fig:color_images} shown the two color images named ``Caps'' and ``Baboon.''

All images in these sets are corrupted by multiplicative noise, generated via a Gamma distribution with a parameter known as the number of looks $L$, sampled from the set $\{1, 3, 5, 10\}$. The number of looks controls the noise intensity, with lower values indicating higher noise. 

\begin{figure}[H]
    \centering
    \begin{subfigure}[b]{0.24\textwidth}
        \centering
        \includegraphics[width=\textwidth]{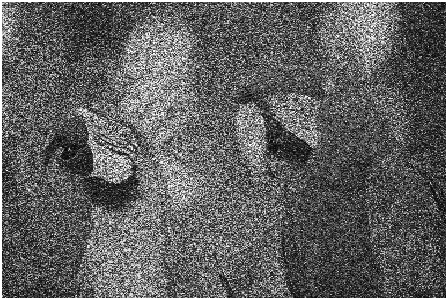}
        \caption{Look=1}
    \end{subfigure}
    \begin{subfigure}[b]{0.24\textwidth}
        \centering
        \includegraphics[width=\textwidth]{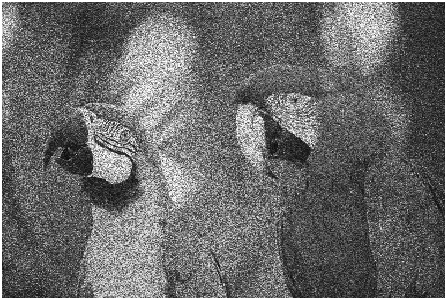}
        \caption{Look=3}
    \end{subfigure}
    \begin{subfigure}[b]{0.24\textwidth}
        \centering
        \includegraphics[width=\textwidth]{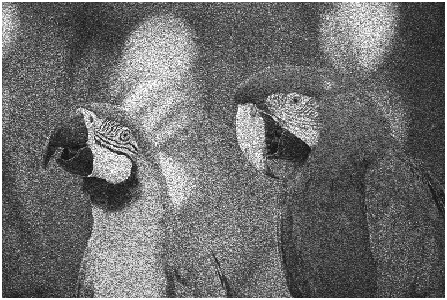}
        \caption{Look=5}
    \end{subfigure}
    \begin{subfigure}[b]{0.24\textwidth}
        \centering
        \includegraphics[width=\textwidth]{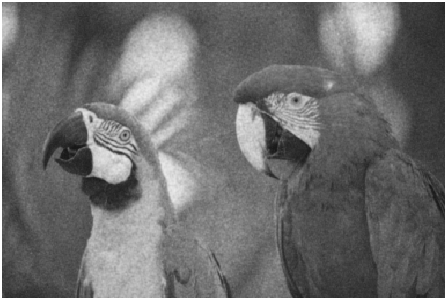}
        \caption{Look=10}
    \end{subfigure}

    \begin{subfigure}[b]{0.24\textwidth}
        \centering
        \includegraphics[width=\textwidth]{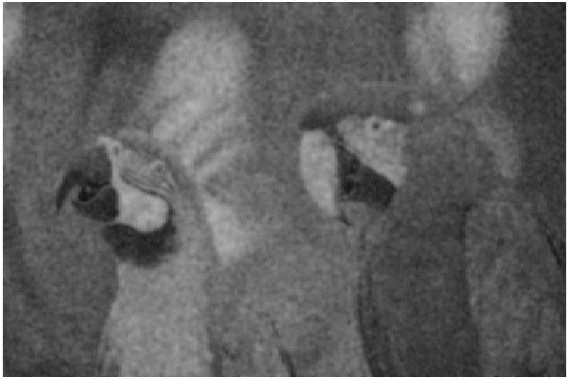}
        \caption{TDM}
    \end{subfigure}
    \begin{subfigure}[b]{0.24\textwidth}
        \centering
        \includegraphics[width=\textwidth]{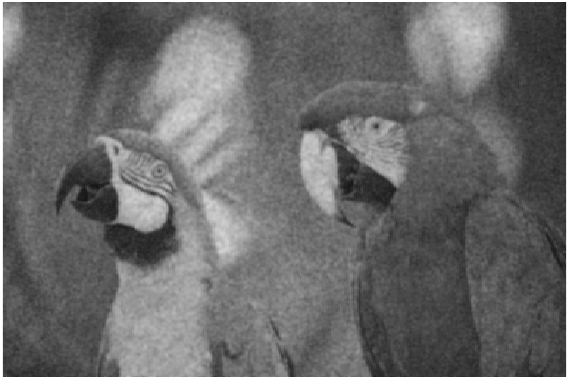}
        \caption{TDM}
    \end{subfigure}
    \begin{subfigure}[b]{0.24\textwidth}
        \centering
        \includegraphics[width=\textwidth]{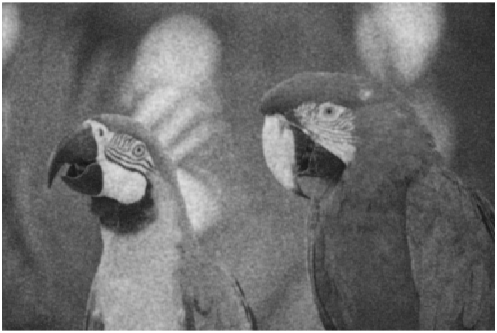}
        \caption{TDM}
    \end{subfigure}
    \begin{subfigure}[b]{0.24\textwidth}
        \centering
        \includegraphics[width=\textwidth]{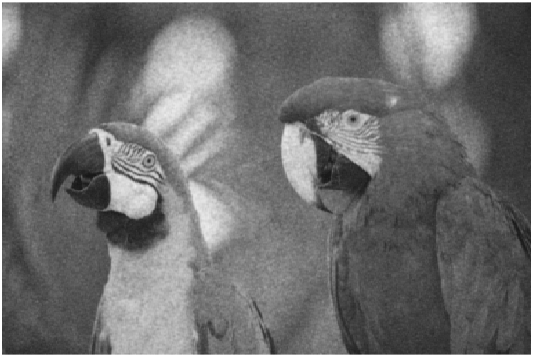}
        \caption{TDM}
    \end{subfigure}

    \begin{subfigure}[b]{0.24\textwidth}
        \centering
        \includegraphics[width=\textwidth]{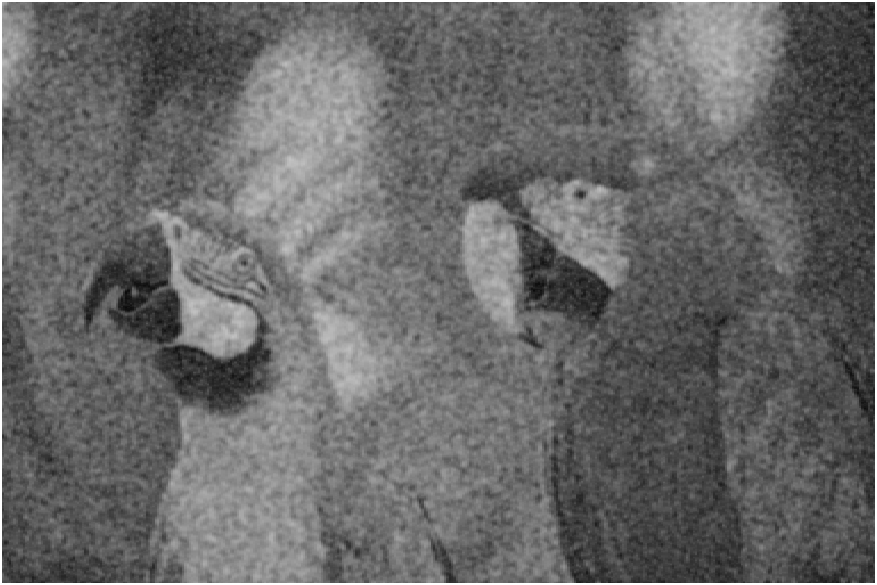}
        \caption{TDFM}
    \end{subfigure}
    \begin{subfigure}[b]{0.24\textwidth}
        \centering
        \includegraphics[width=\textwidth]{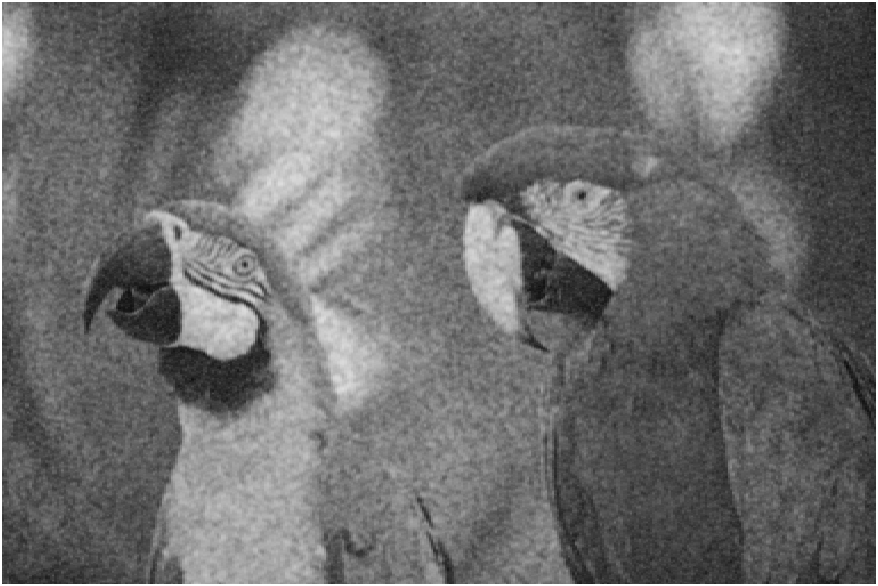}
        \caption{TDFM}
    \end{subfigure}
    \begin{subfigure}[b]{0.24\textwidth}
        \centering
        \includegraphics[width=\textwidth]{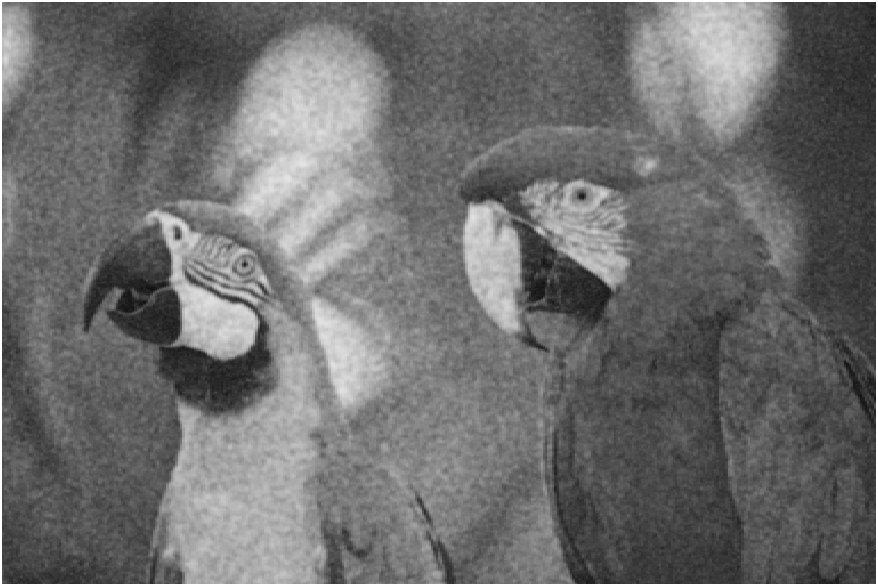}
        \caption{ TDFM}
    \end{subfigure}
    \begin{subfigure}[b]{0.24\textwidth}
        \centering
        \includegraphics[width=\textwidth]{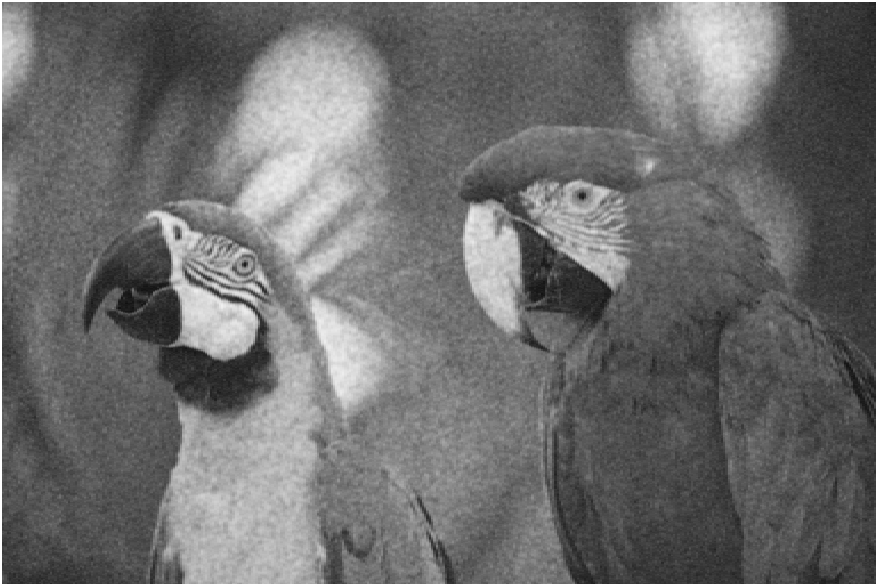}
        \caption{TDFM}
    \end{subfigure}

    \begin{subfigure}[b]{0.24\textwidth}
        \centering
        \includegraphics[width=\textwidth]{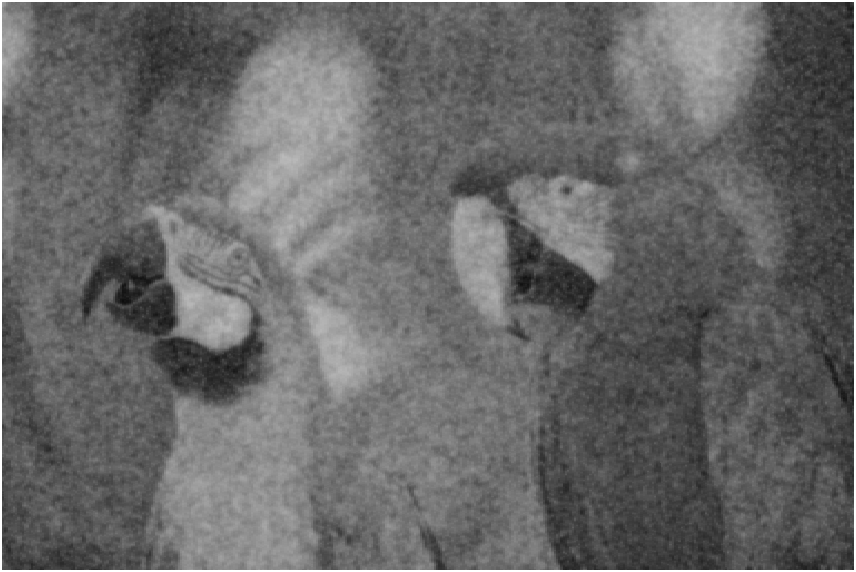}
        \caption{Model 2}
    \end{subfigure}
    \begin{subfigure}[b]{0.24\textwidth}
        \centering
        \includegraphics[width=\textwidth]{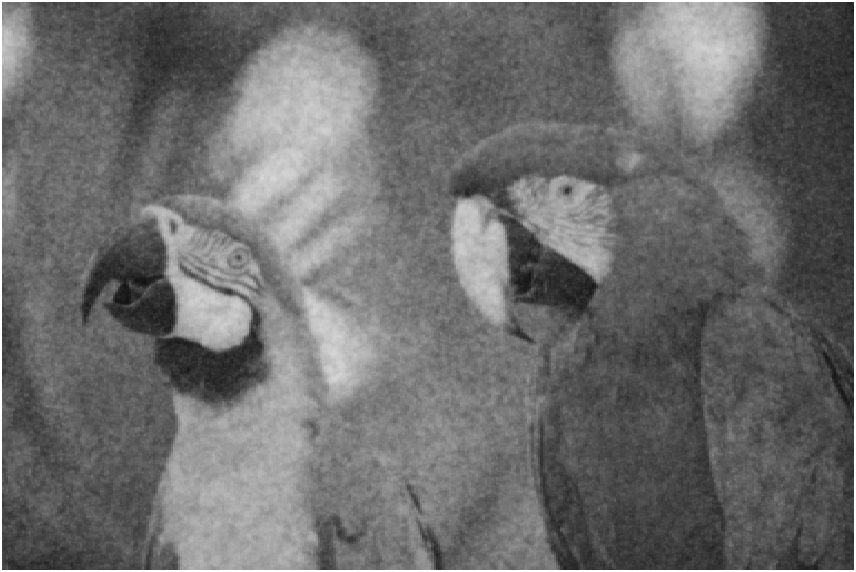}
        \caption{Model 2}
    \end{subfigure}
    \begin{subfigure}[b]{0.24\textwidth}
        \centering
        \includegraphics[width=\textwidth]{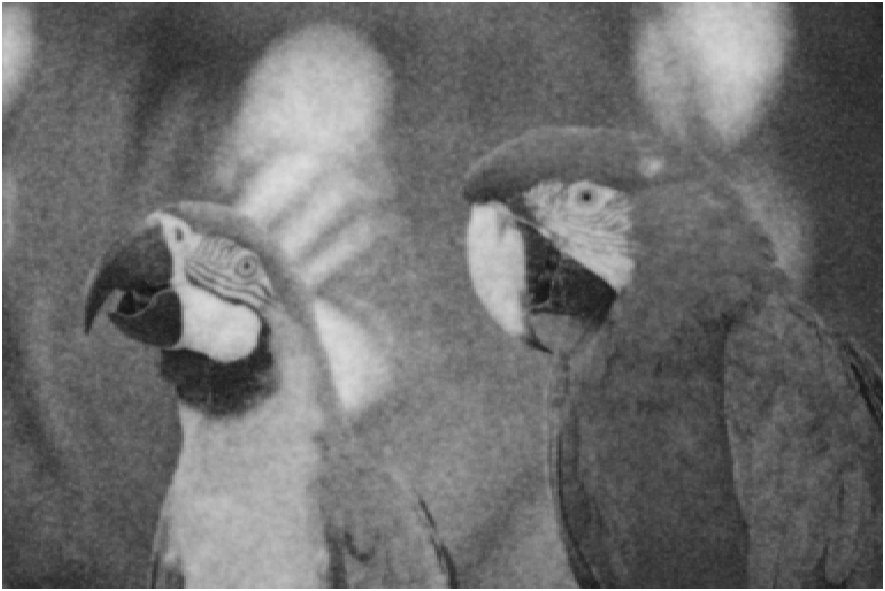}
        \caption{Model 2}
    \end{subfigure}
    \begin{subfigure}[b]{0.24\textwidth}
        \centering
        \includegraphics[width=\textwidth]{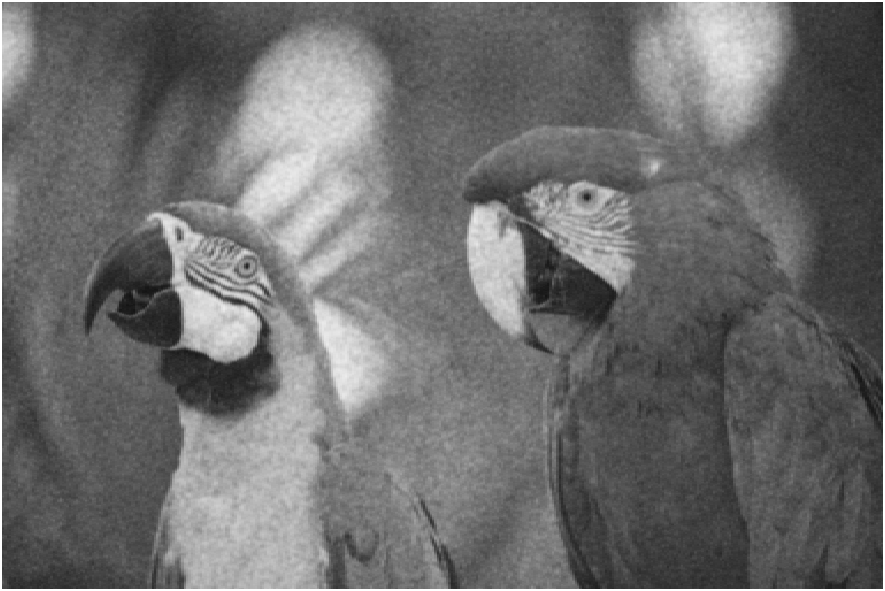}
        \caption{Model 2}
    \end{subfigure}
    \begin{subfigure}[b]{0.24\textwidth}
        \centering
        \includegraphics[width=\textwidth]{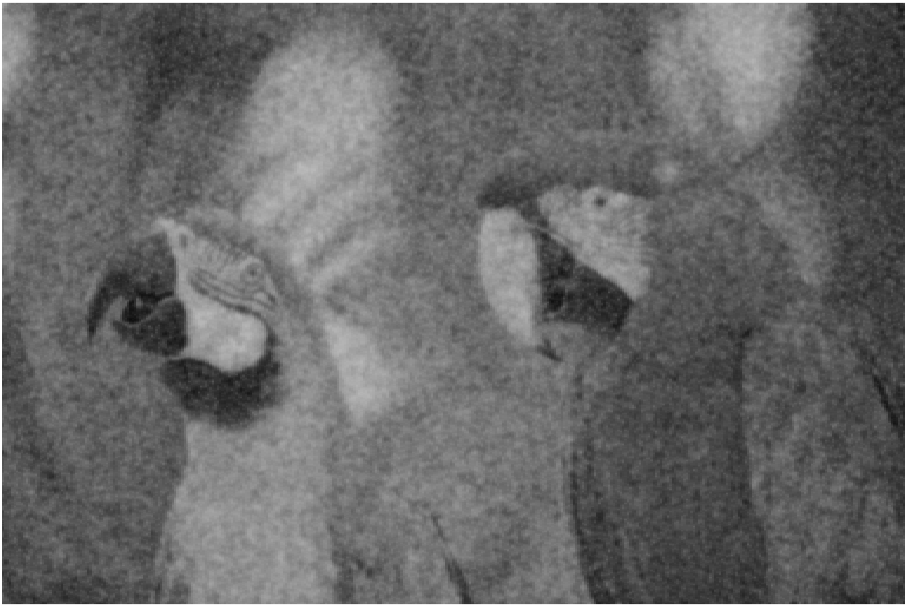}
        \caption{Model 1}
    \end{subfigure}
    \begin{subfigure}[b]{0.24\textwidth}
        \centering
        \includegraphics[width=\textwidth]{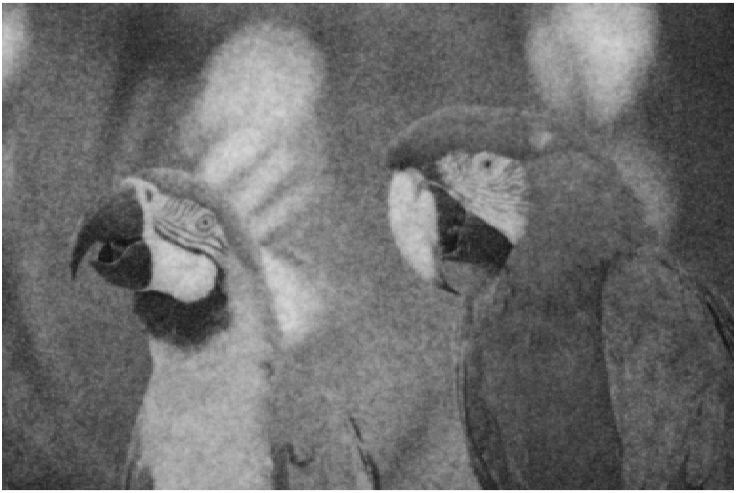}
        \caption{Model 1}
    \end{subfigure}
    \begin{subfigure}[b]{0.24\textwidth}
        \centering
        \includegraphics[width=\textwidth]{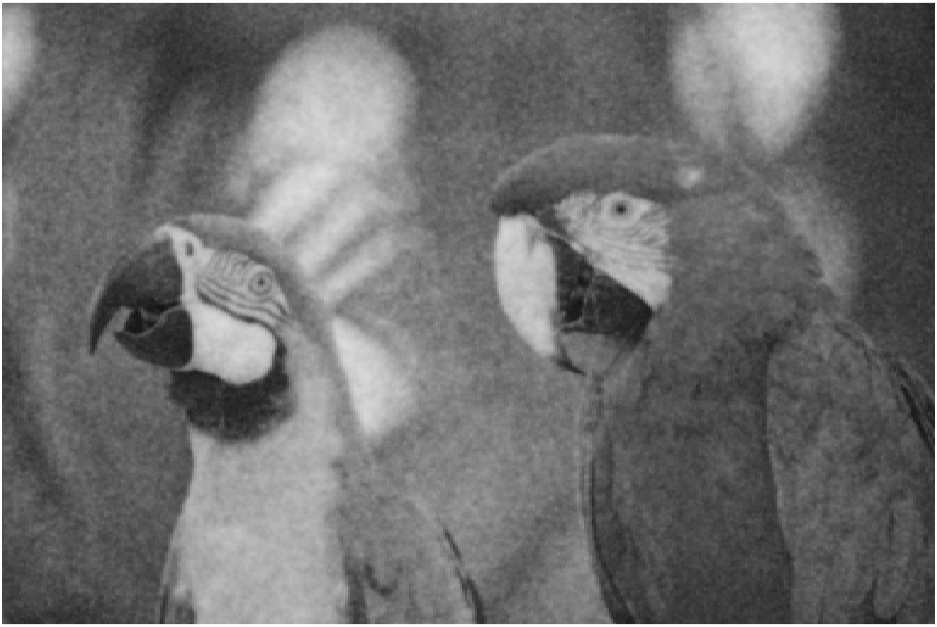}
        \caption{Model 1}
    \end{subfigure}
    \begin{subfigure}[b]{0.24\textwidth}
        \centering
        \includegraphics[width=\textwidth]{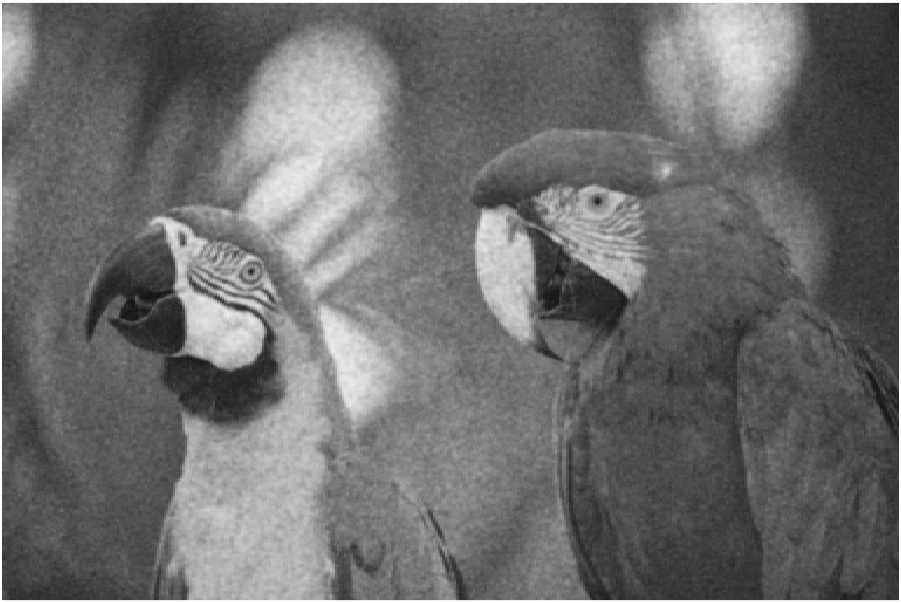}
        \caption{Model 1}
    \end{subfigure}
    \caption{
     The first row contains noisy parrot images with noise level Look = 1, 3, 5, 10. 
        Subsequent rows: Restored parrot images using different models (TDM, TDFM, Model 2, Model 1).
    }
    \label{fig:gamma_noise_results_parrots}
\end{figure}

\begin{figure}[H]
    \centering
    \begin{subfigure}[b]{0.24\textwidth}
        \centering
        \includegraphics[width=\textwidth]{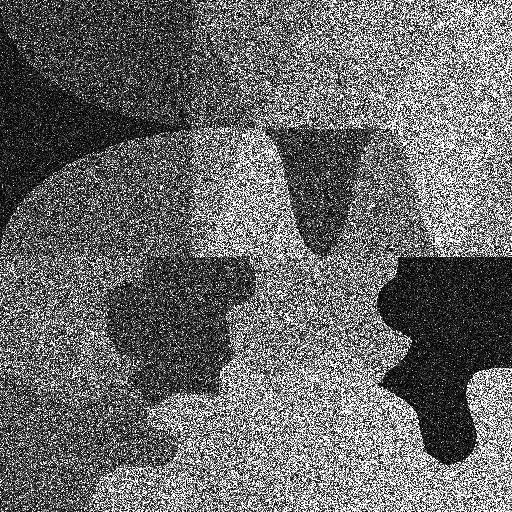}
        \caption{Look=1}
    \end{subfigure}
    \begin{subfigure}[b]{0.24\textwidth}
        \centering
        \includegraphics[width=\textwidth]{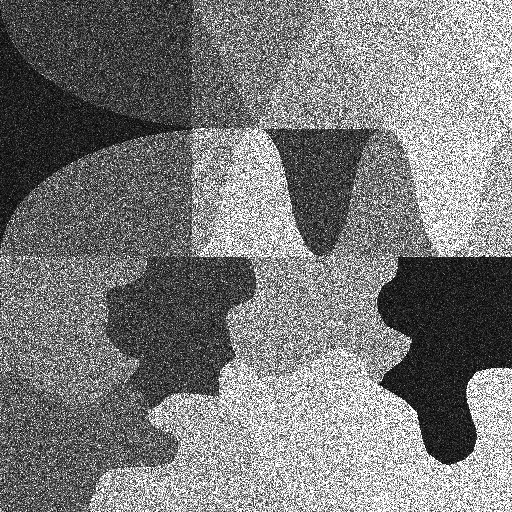}
        \caption{Look=3}
    \end{subfigure}
    \begin{subfigure}[b]{0.24\textwidth}
        \centering
        \includegraphics[width=\textwidth]{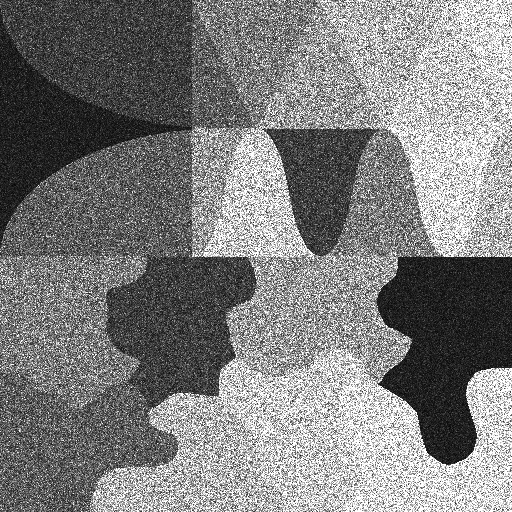}
        \caption{Look=5}
    \end{subfigure}
    \begin{subfigure}[b]{0.24\textwidth}
        \centering
        \includegraphics[width=\textwidth]{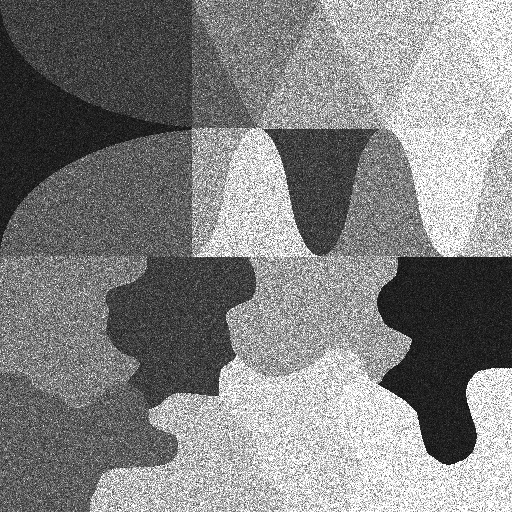}
        \caption{Look=10}
    \end{subfigure}

    \begin{subfigure}[b]{0.24\textwidth}
        \centering
        \includegraphics[width=\textwidth]{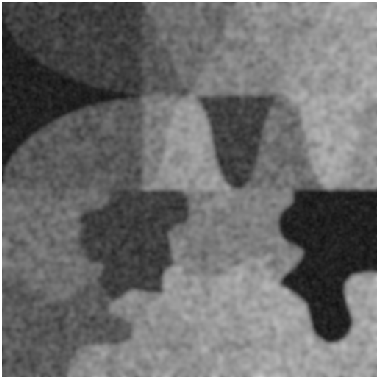}
        \caption{TDM}
    \end{subfigure}
    \begin{subfigure}[b]{0.24\textwidth}
        \centering
        \includegraphics[width=\textwidth]{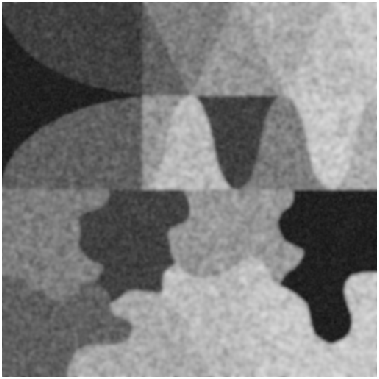}
        \caption{TDM}
    \end{subfigure}
    \begin{subfigure}[b]{0.24\textwidth}
        \centering
        \includegraphics[width=\textwidth]{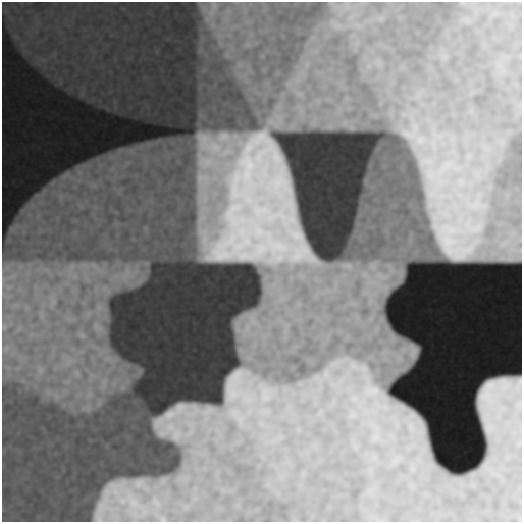}
        \caption{TDM}
    \end{subfigure}
    \begin{subfigure}[b]{0.24\textwidth}
        \centering
        \includegraphics[width=\textwidth]{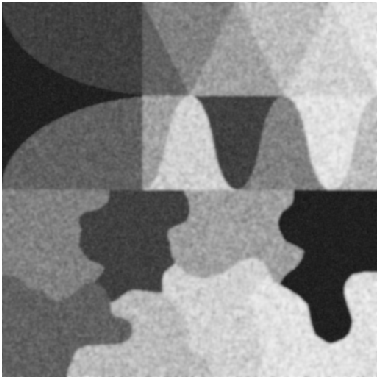}
        \caption{TDM}
    \end{subfigure}

    \begin{subfigure}[b]{0.24\textwidth}
        \centering
        \includegraphics[width=\textwidth]{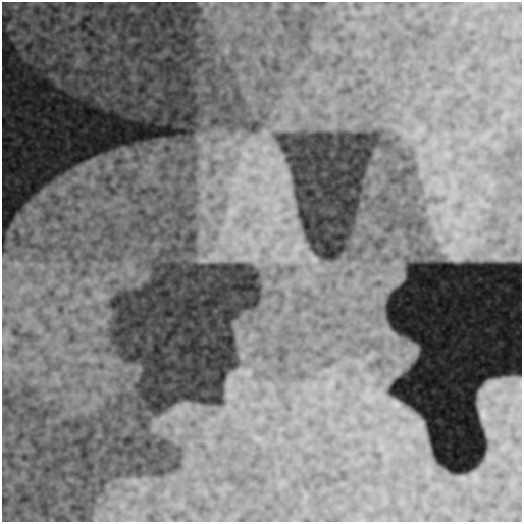}
        \caption{TDFM}
    \end{subfigure}
    \begin{subfigure}[b]{0.24\textwidth}
        \centering
        \includegraphics[width=\textwidth]{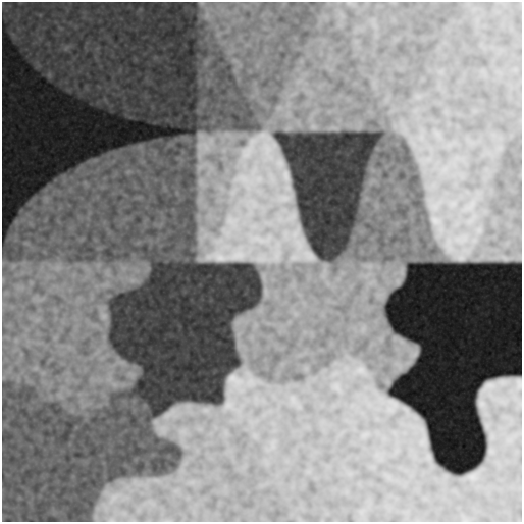 }
        \caption{TDFM}
    \end{subfigure}
    \begin{subfigure}[b]{0.24\textwidth}
        \centering
        \includegraphics[width=\textwidth]{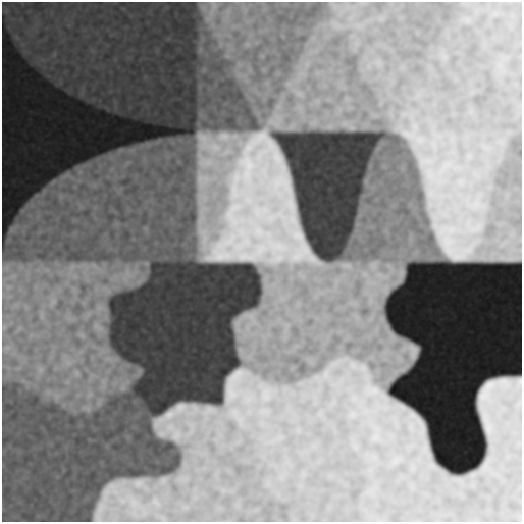}
        \caption{ TDFM}
    \end{subfigure}
    \begin{subfigure}[b]{0.24\textwidth}
        \centering
        \includegraphics[width=\textwidth]{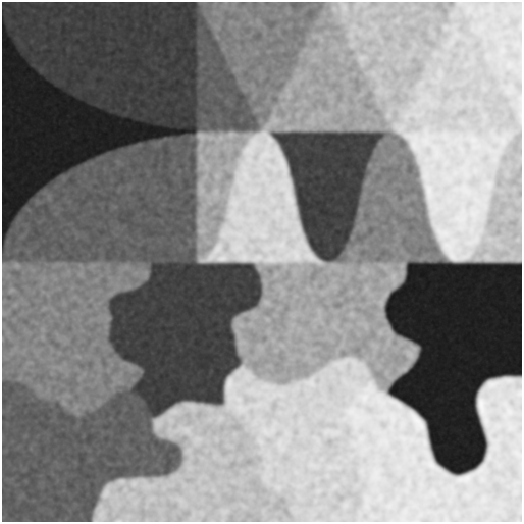}
        \caption{TDFM}
    \end{subfigure}

    \begin{subfigure}[b]{0.24\textwidth}
        \centering
        \includegraphics[width=\textwidth]{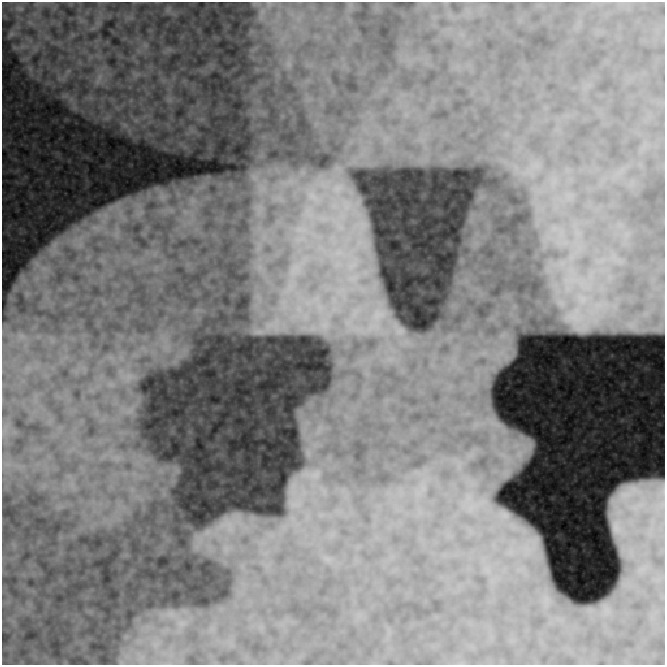 }
        \caption{Model 2}
    \end{subfigure}
    \begin{subfigure}[b]{0.24\textwidth}
        \centering
        \includegraphics[width=\textwidth]{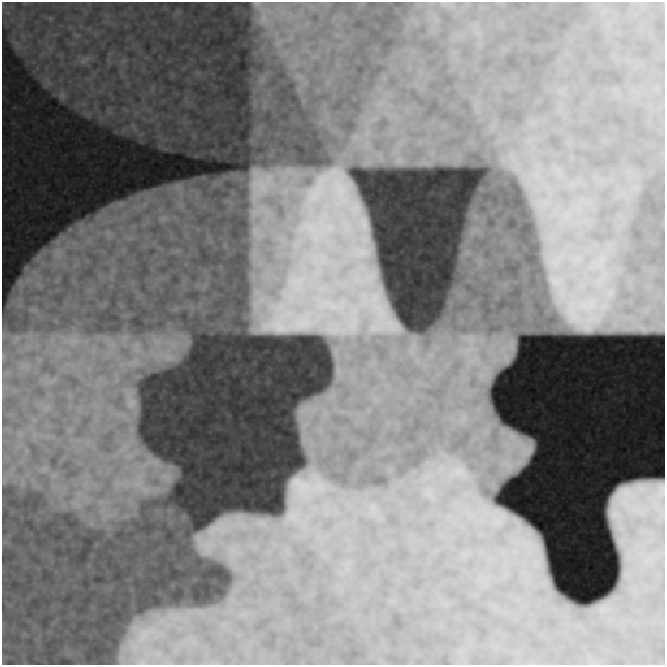}
        \caption{Model 2}
    \end{subfigure}
    \begin{subfigure}[b]{0.24\textwidth}
        \centering
        \includegraphics[width=\textwidth]{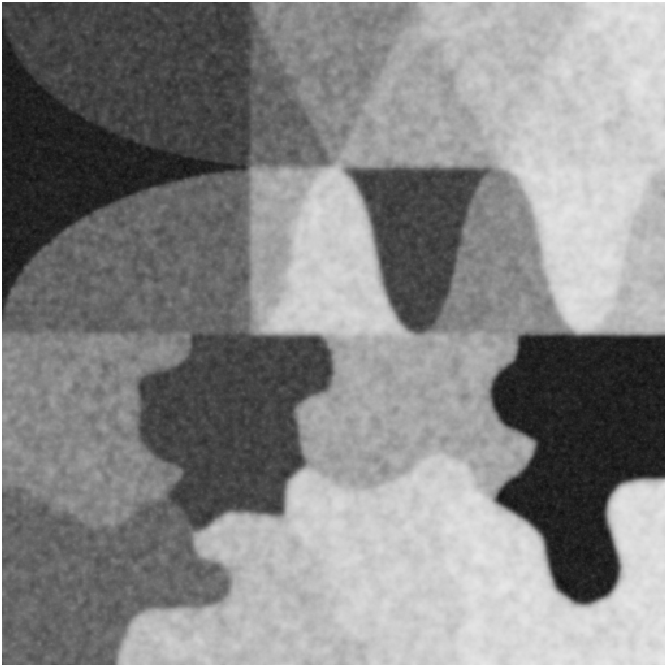}
        \caption{Model 2}
    \end{subfigure}
    \begin{subfigure}[b]{0.24\textwidth}
        \centering
        \includegraphics[width=\textwidth]{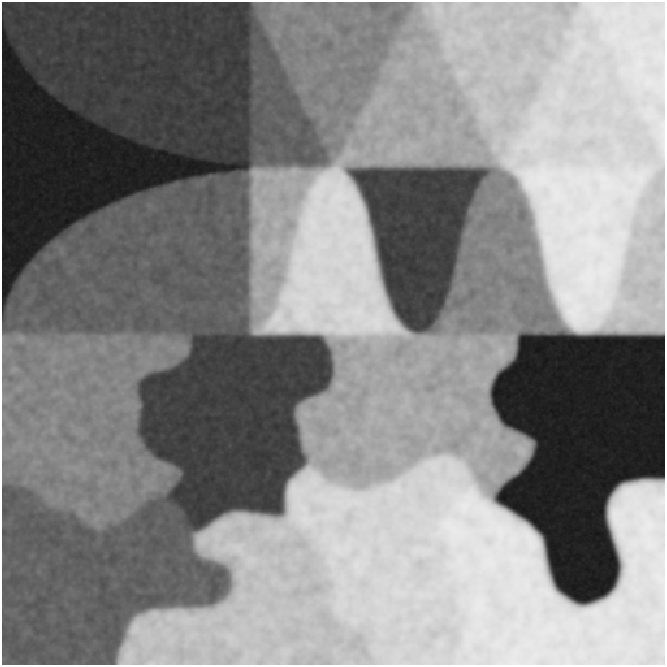}
        \caption{Model 2}
    \end{subfigure}
    \begin{subfigure}[b]{0.24\textwidth}
        \centering
        \includegraphics[width=\textwidth]{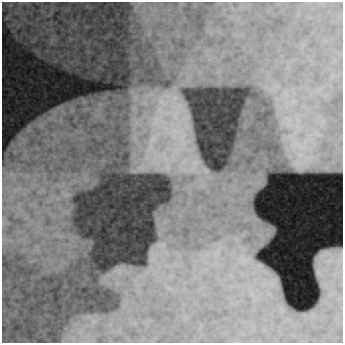 }
        \caption{Model 1}
    \end{subfigure}
    \begin{subfigure}[b]{0.24\textwidth}
        \centering
        \includegraphics[width=\textwidth]{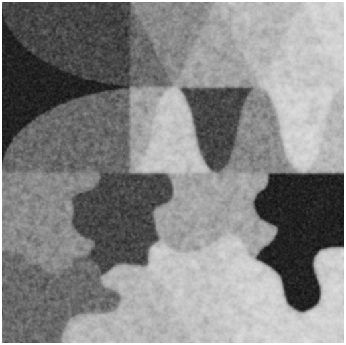}
        \caption{Model 1}
    \end{subfigure}
    \begin{subfigure}[b]{0.24\textwidth}
        \centering
        \includegraphics[width=\textwidth]{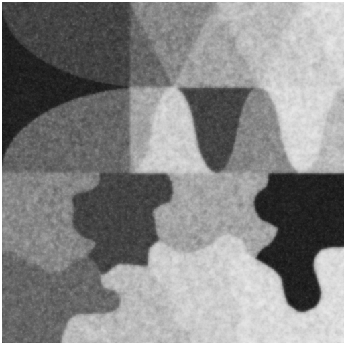}
        \caption{Model 1}
    \end{subfigure}
    \begin{subfigure}[b]{0.24\textwidth}
        \centering
        \includegraphics[width=\textwidth]{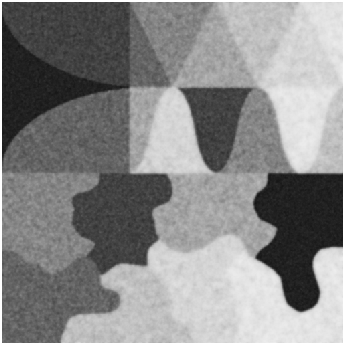}
        \caption{Model 1}
    \end{subfigure}
    \caption{
     The first row contains noisy texture images with noise level Look = 1, 3, 5, 10. 
        Subsequent rows: Restored texture images using different models (TDM, TDFM, Model 2, Model 1).
    }
    \label{fig:gamma_noise_results_texture}
\end{figure}

\begin{figure}[H]
    \centering
    \begin{subfigure}[b]{0.24\textwidth}
        \centering
        \includegraphics[width=\textwidth]{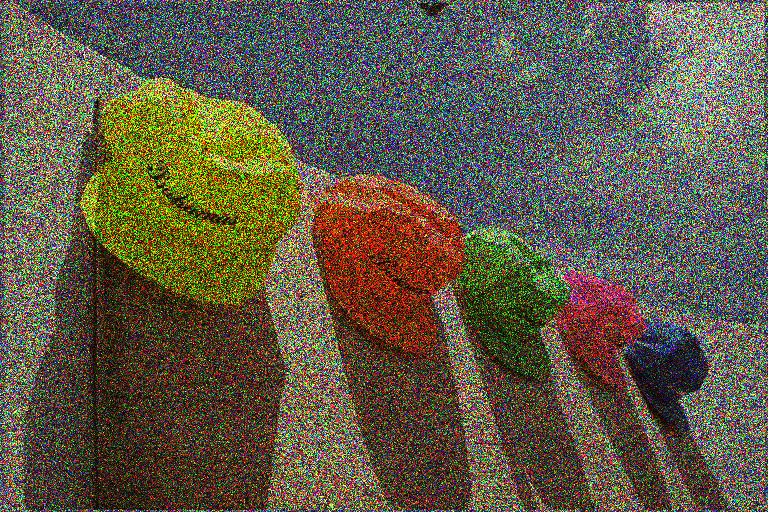}
        \caption{Look=1}
    \end{subfigure}
    \begin{subfigure}[b]{0.24\textwidth}
        \centering
        \includegraphics[width=\textwidth]{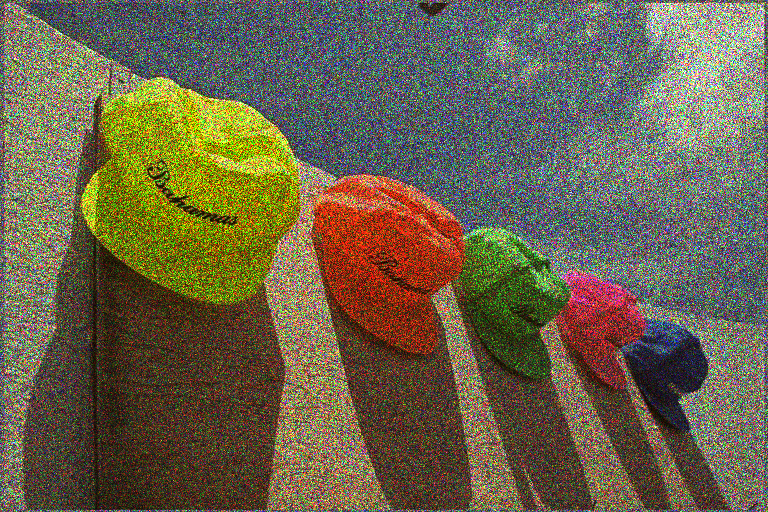}
        \caption{Look=3}
    \end{subfigure}
    \begin{subfigure}[b]{0.24\textwidth}
        \centering
        \includegraphics[width=\textwidth]{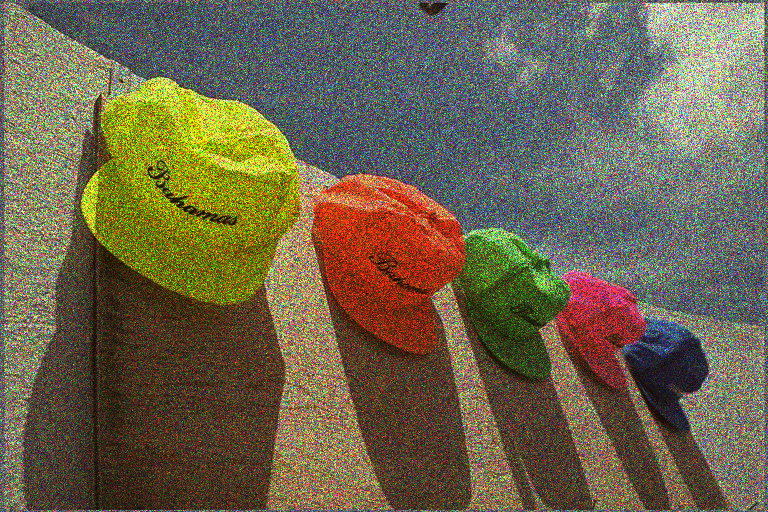}
        \caption{Look=5}
    \end{subfigure}
    \begin{subfigure}[b]{0.24\textwidth}
        \centering
        \includegraphics[width=\textwidth]{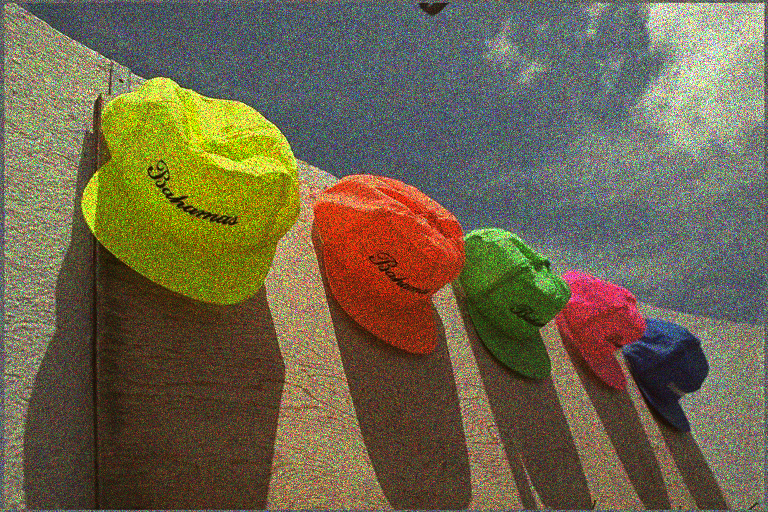}
        \caption{Look=10}
    \end{subfigure}

    \begin{subfigure}[b]{0.24\textwidth}
        \centering
        \includegraphics[width=\textwidth]{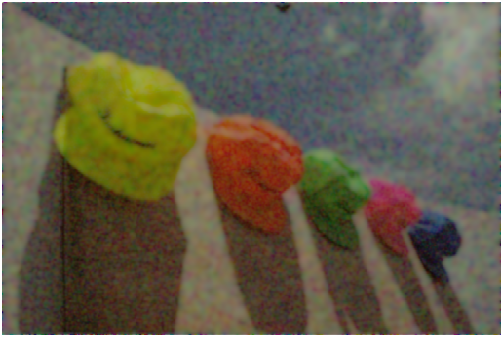}
        \caption{TDM}
    \end{subfigure}
    \begin{subfigure}[b]{0.24\textwidth}
        \centering
        \includegraphics[width=\textwidth]{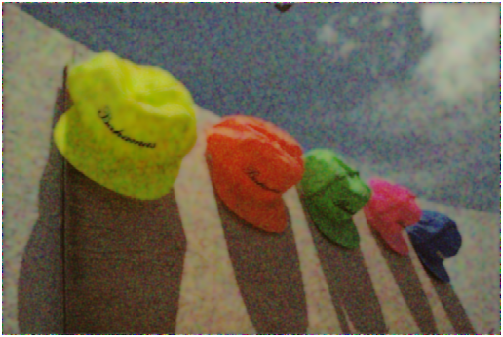}
        \caption{TDM}
    \end{subfigure}
    \begin{subfigure}[b]{0.24\textwidth}
        \centering
        \includegraphics[width=\textwidth]{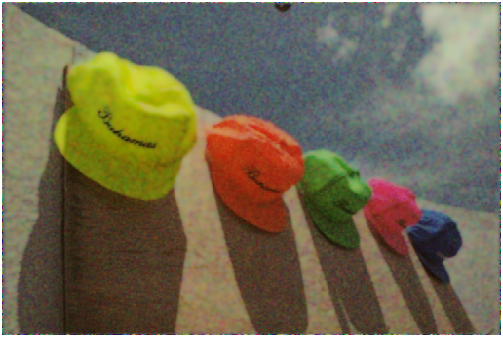}
        \caption{TDM}
    \end{subfigure}
    \begin{subfigure}[b]{0.24\textwidth}
        \centering
        \includegraphics[width=\textwidth]{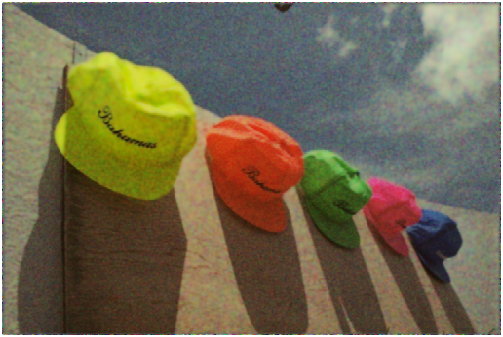}
        \caption{TDM}
    \end{subfigure}

    \begin{subfigure}[b]{0.24\textwidth}
        \centering
        \includegraphics[width=\textwidth]{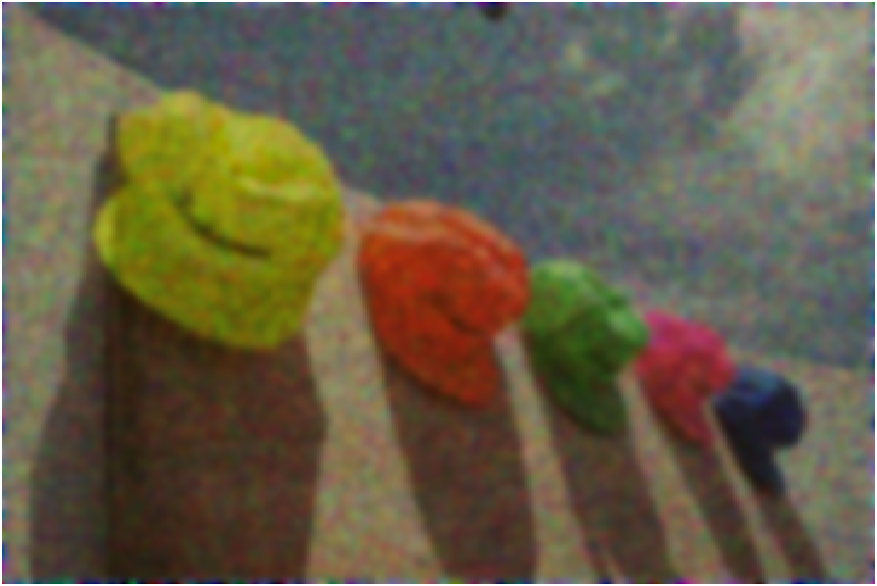}
        \caption{TDFM}
    \end{subfigure}
    \begin{subfigure}[b]{0.24\textwidth}
        \centering
        \includegraphics[width=\textwidth]{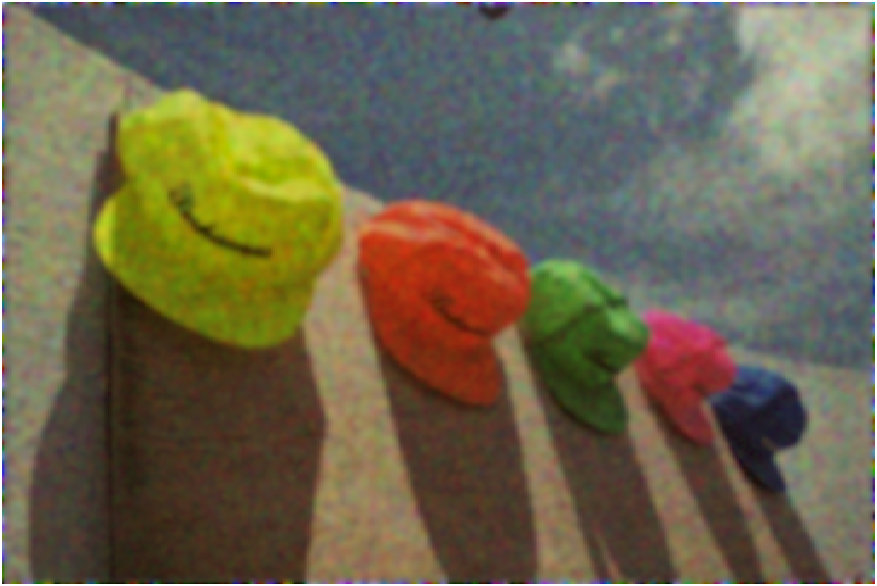}
        \caption{TDFM}
    \end{subfigure}
    \begin{subfigure}[b]{0.24\textwidth}
        \centering
        \includegraphics[width=\textwidth]{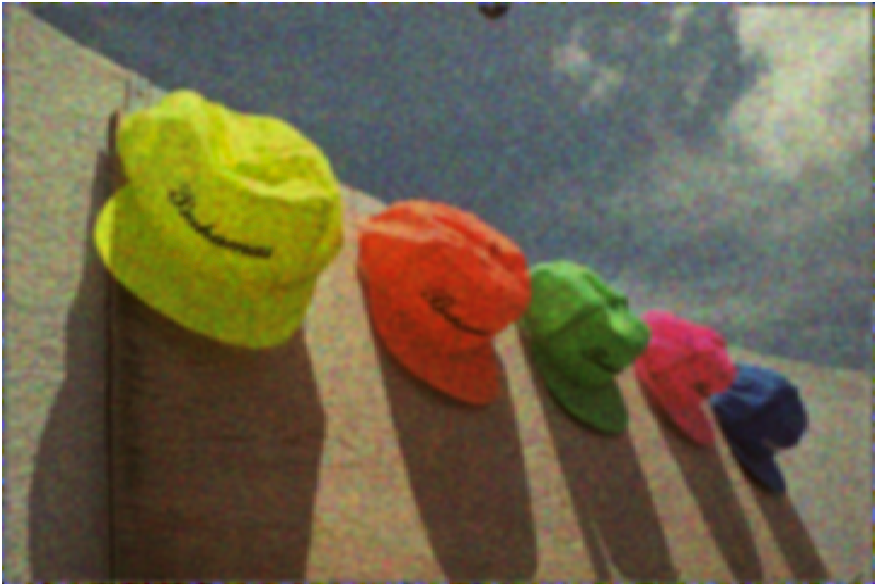}
        \caption{ TDFM}
    \end{subfigure}
    \begin{subfigure}[b]{0.24\textwidth}
        \centering
        \includegraphics[width=\textwidth]{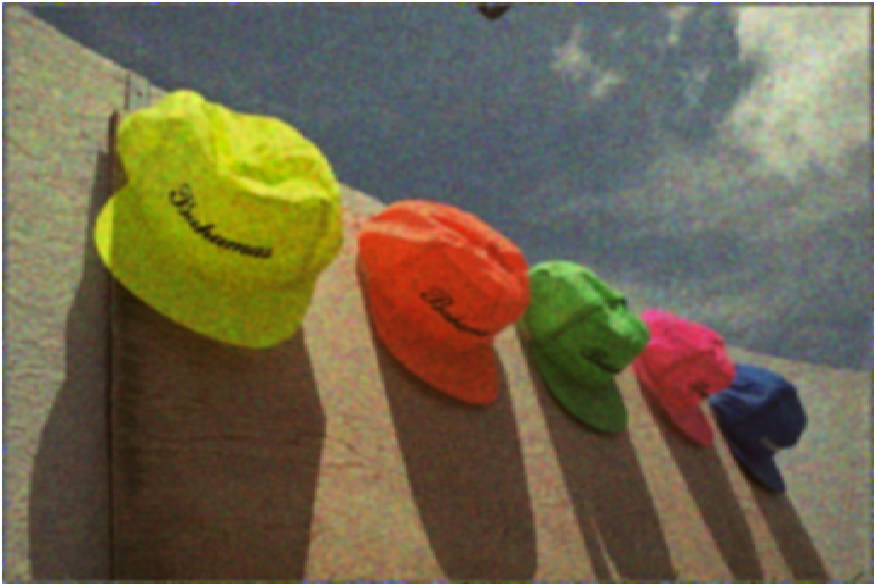}
        \caption{TDFM}
    \end{subfigure}

    \begin{subfigure}[b]{0.24\textwidth}
        \centering
        \includegraphics[width=\textwidth]{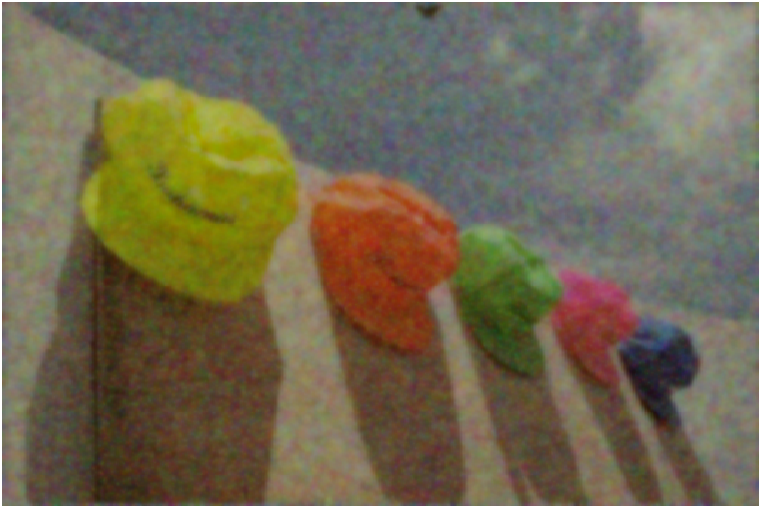 }
        \caption{Model 2}
    \end{subfigure}
    \begin{subfigure}[b]{0.24\textwidth}
        \centering
        \includegraphics[width=\textwidth]{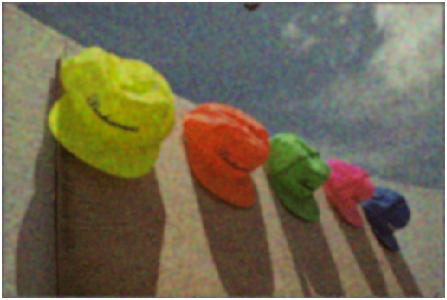}
        \caption{Model 2}
    \end{subfigure}
    \begin{subfigure}[b]{0.24\textwidth}
        \centering
        \includegraphics[width=\textwidth]{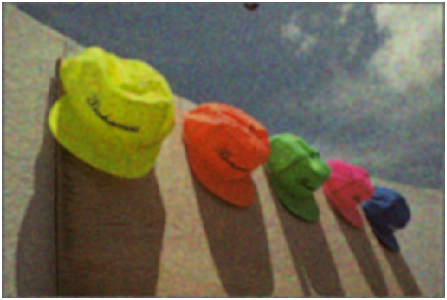}
        \caption{Model 2}
    \end{subfigure}
    \begin{subfigure}[b]{0.24\textwidth}
        \centering
        \includegraphics[width=\textwidth]{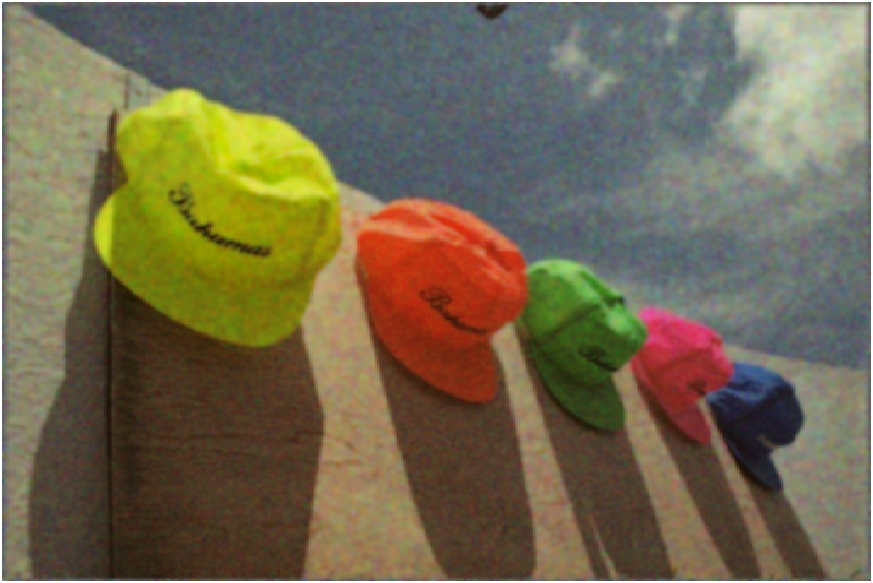}
        \caption{Model 2}
    \end{subfigure}
    \begin{subfigure}[b]{0.24\textwidth}
        \centering
        \includegraphics[width=\textwidth]{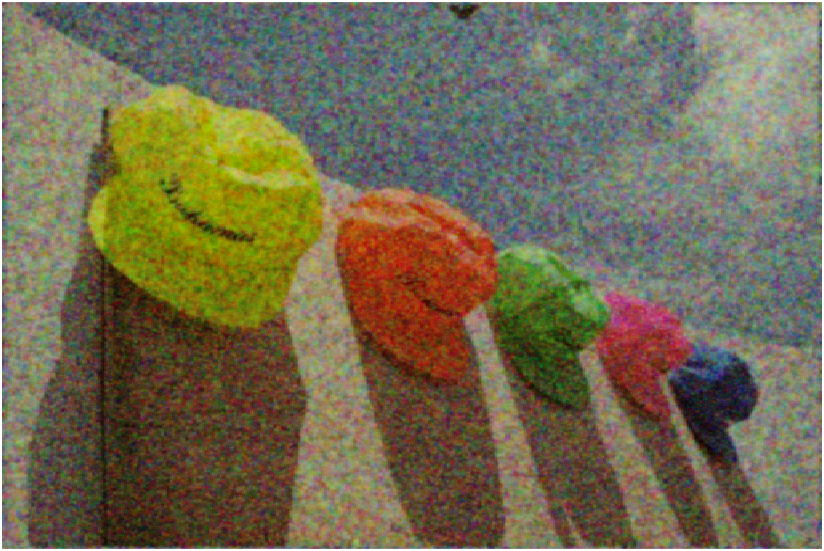 }
        \caption{Model 1}
    \end{subfigure}
    \begin{subfigure}[b]{0.24\textwidth}
        \centering
        \includegraphics[width=\textwidth]{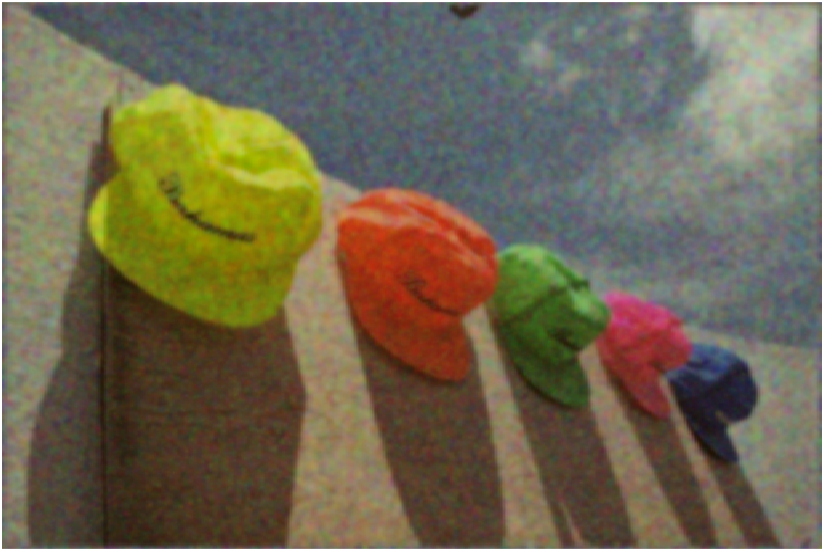}
        \caption{Model 1}
    \end{subfigure}
    \begin{subfigure}[b]{0.24\textwidth}
        \centering
        \includegraphics[width=\textwidth]{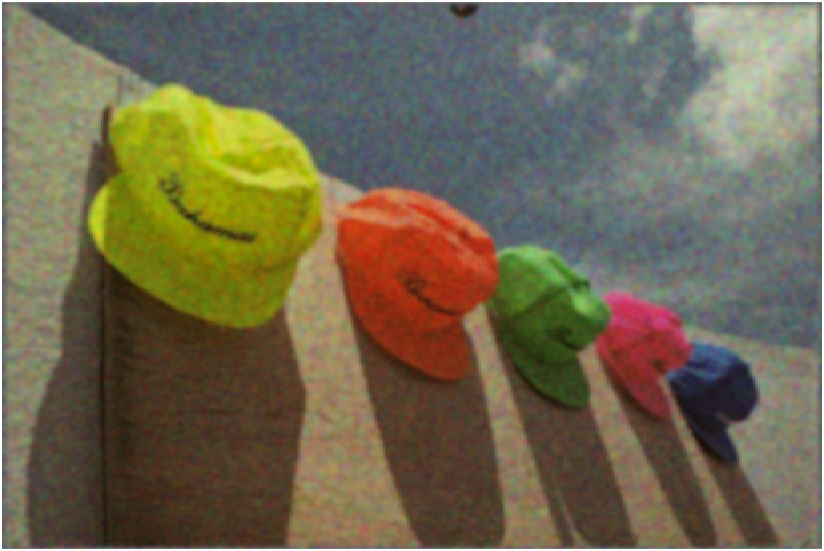}
        \caption{Model 1}
    \end{subfigure}
    \begin{subfigure}[b]{0.24\textwidth}
        \centering
        \includegraphics[width=\textwidth]{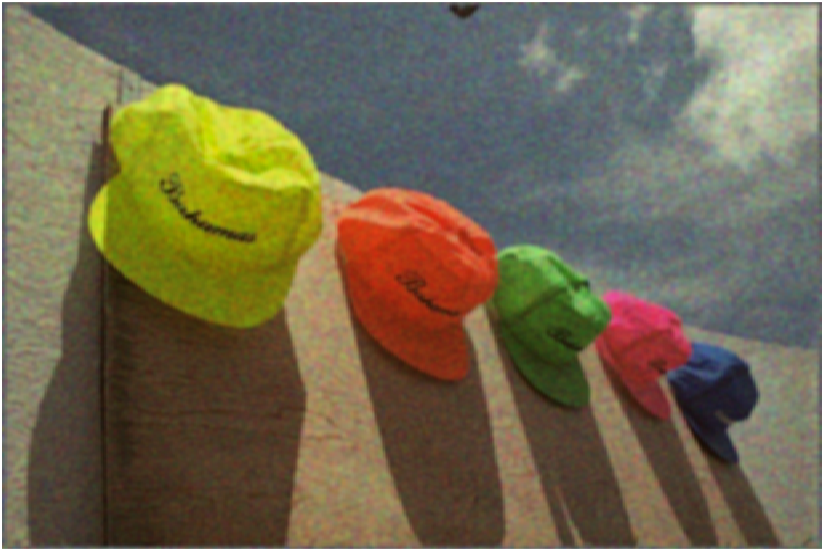}
        \caption{Model 1}
    \end{subfigure}
    \caption{
     The first row contains noisy caps images with noise level Look = 1, 3, 5, 10. 
        Subsequent rows: Restored caps images using different models (TDM, TDFM, Model 2, Model 1).
    }
    \label{fig:gamma_noise_results_caps}
\end{figure}

\begin{figure}[H]
    \centering
    \begin{subfigure}[b]{0.24\textwidth}
        \centering
        \includegraphics[width=\textwidth]{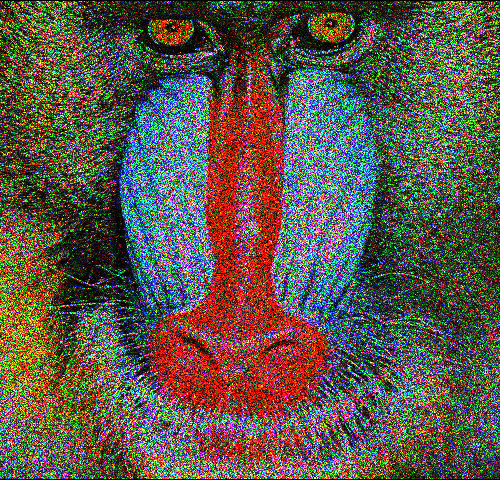}
        \caption{Look=1}
    \end{subfigure}
    \begin{subfigure}[b]{0.24\textwidth}
        \centering
        \includegraphics[width=\textwidth]{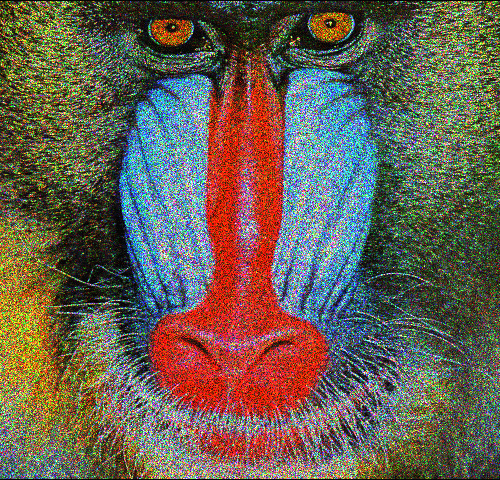}
        \caption{Look=3}
    \end{subfigure}
    \begin{subfigure}[b]{0.24\textwidth}
        \centering
        \includegraphics[width=\textwidth]{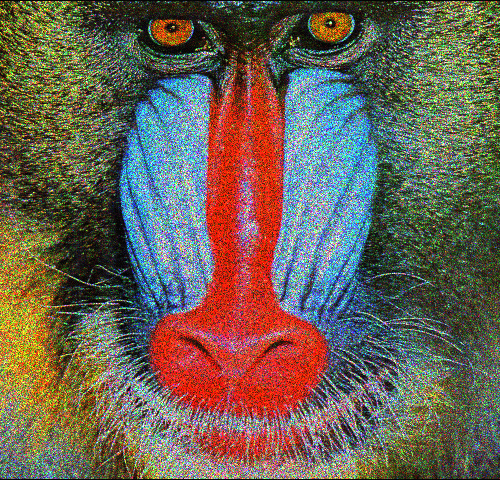}
        \caption{Look=5}
    \end{subfigure}
    \begin{subfigure}[b]{0.24\textwidth}
        \centering
        \includegraphics[width=\textwidth]{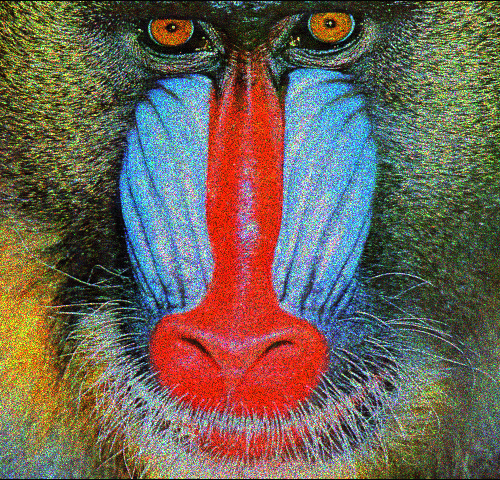}
        \caption{Look=10}
    \end{subfigure}

    \begin{subfigure}[b]{0.24\textwidth}
        \centering
        \includegraphics[width=\textwidth]{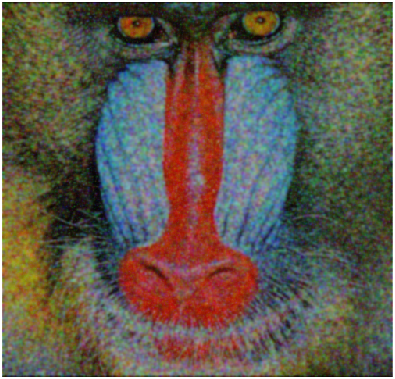}
        \caption{TDM}
    \end{subfigure}
    \begin{subfigure}[b]{0.24\textwidth}
        \centering
        \includegraphics[width=\textwidth]{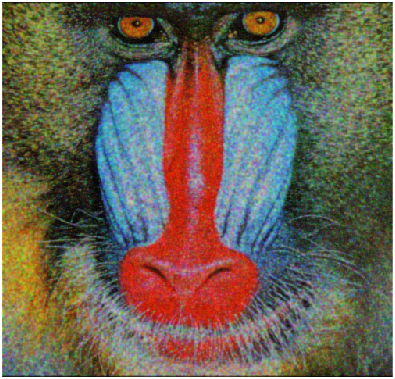}
        \caption{TDM}
    \end{subfigure}
    \begin{subfigure}[b]{0.24\textwidth}
        \centering
        \includegraphics[width=\textwidth]{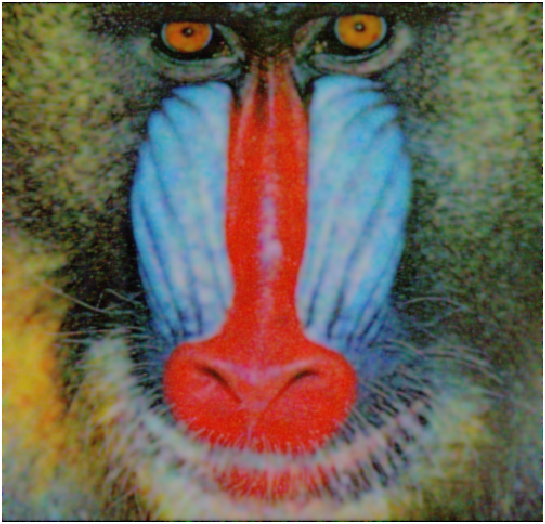}
        \caption{TDM}
    \end{subfigure}
    \begin{subfigure}[b]{0.24\textwidth}
        \centering
        \includegraphics[width=\textwidth]{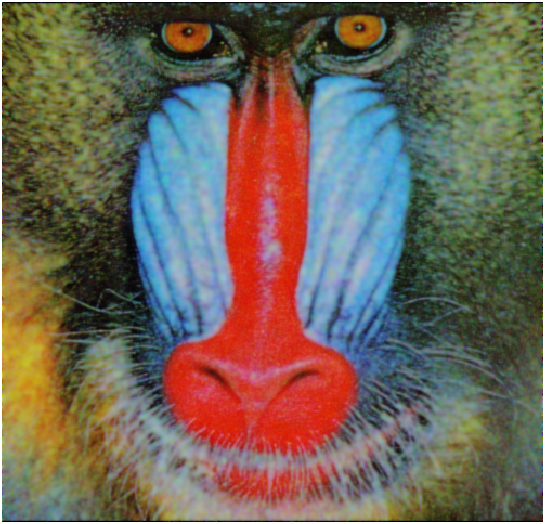}
        \caption{TDM}
    \end{subfigure}

    \begin{subfigure}[b]{0.24\textwidth}
        \centering
        \includegraphics[width=\textwidth]{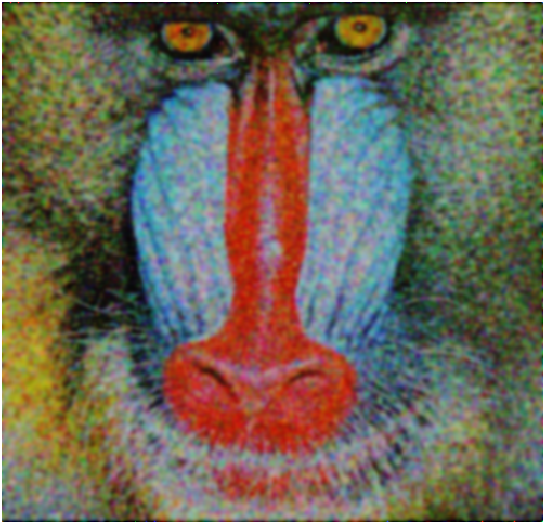}
        \caption{TDFM}
    \end{subfigure}
    \begin{subfigure}[b]{0.24\textwidth}
        \centering
        \includegraphics[width=\textwidth]{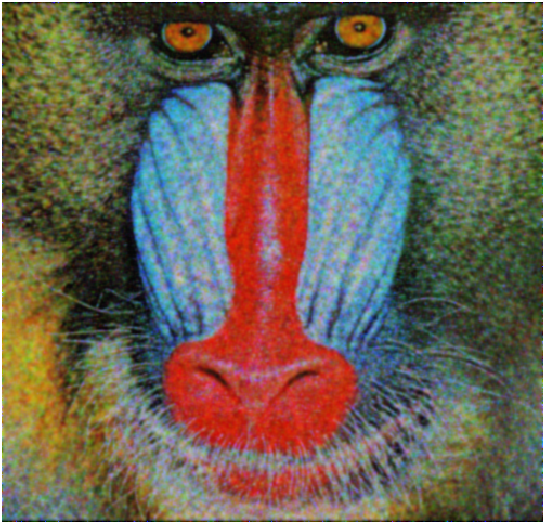}
        \caption{TDFM}
    \end{subfigure}
    \begin{subfigure}[b]{0.24\textwidth}
        \centering
        \includegraphics[width=\textwidth]{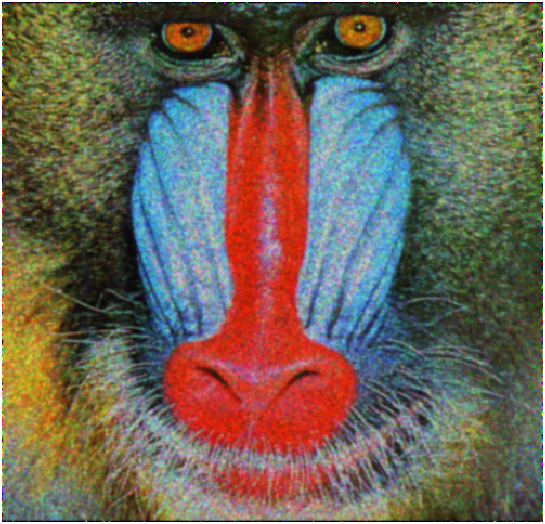}
        \caption{ TDFM}
    \end{subfigure}
    \begin{subfigure}[b]{0.24\textwidth}
        \centering
        \includegraphics[width=\textwidth]{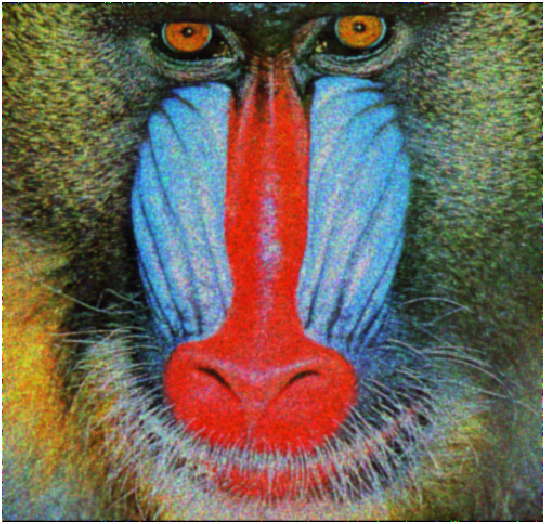}
        \caption{TDFM}
    \end{subfigure}

    \begin{subfigure}[b]{0.24\textwidth}
        \centering
        \includegraphics[width=\textwidth]{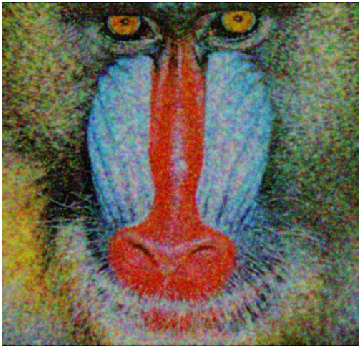 }
        \caption{Model 2}
    \end{subfigure}
    \begin{subfigure}[b]{0.24\textwidth}
        \centering
        \includegraphics[width=\textwidth]{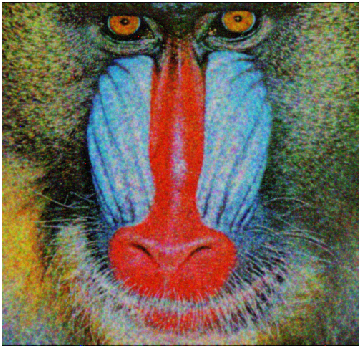}
        \caption{Model 2}
    \end{subfigure}
    \begin{subfigure}[b]{0.24\textwidth}
        \centering
        \includegraphics[width=\textwidth]{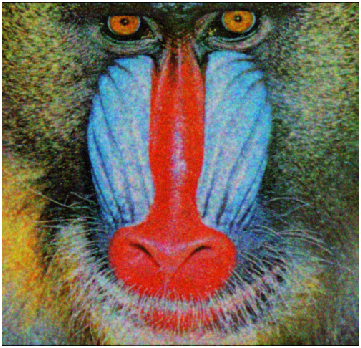}
        \caption{Model 2}
    \end{subfigure}
    \begin{subfigure}[b]{0.24\textwidth}
        \centering
        \includegraphics[width=\textwidth]{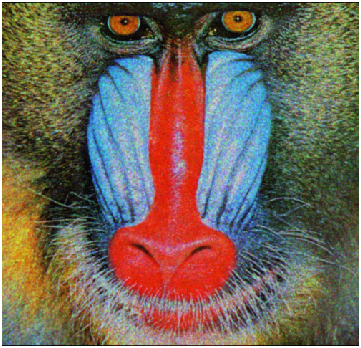}
        \caption{Model 2}
    \end{subfigure}
    \begin{subfigure}[b]{0.24\textwidth}
        \centering
        \includegraphics[width=\textwidth]{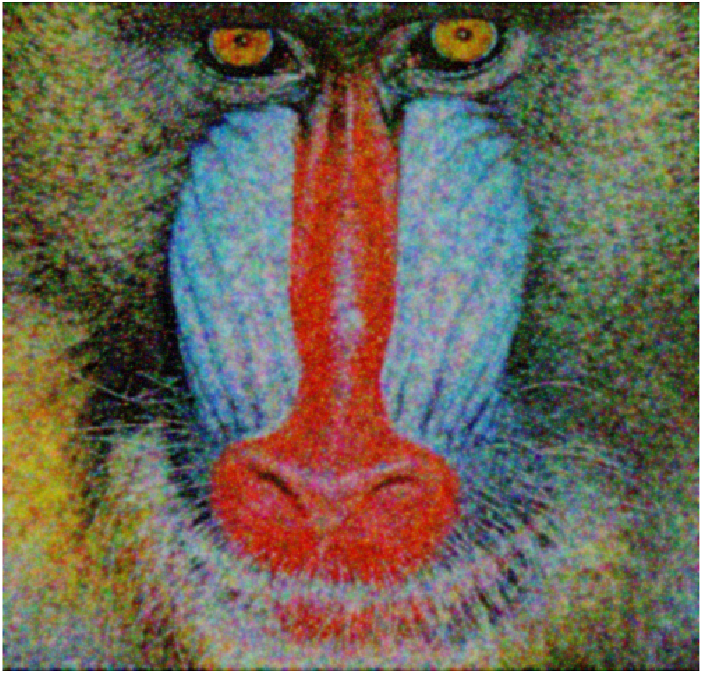 }
        \caption{Model 1}
    \end{subfigure}
    \begin{subfigure}[b]{0.24\textwidth}
        \centering
        \includegraphics[width=\textwidth]{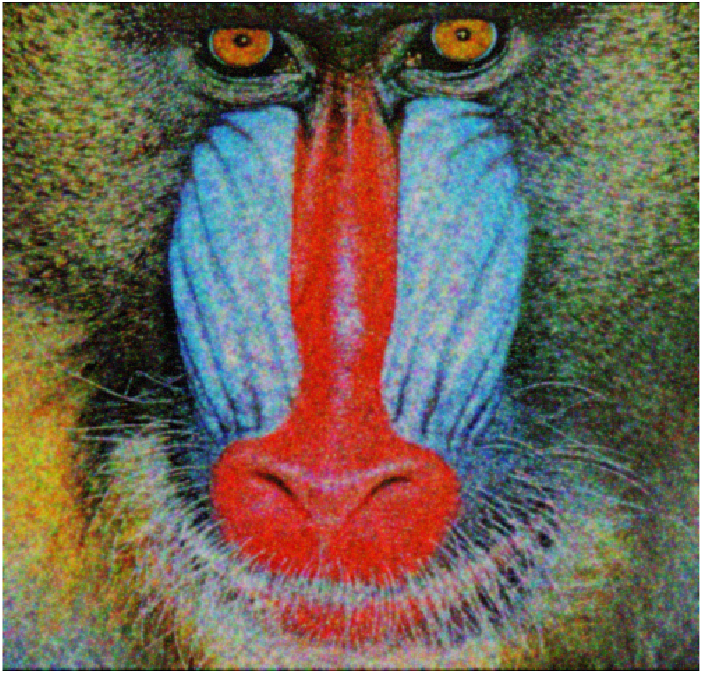}
        \caption{Model 1}
    \end{subfigure}
    \begin{subfigure}[b]{0.24\textwidth}
        \centering
        \includegraphics[width=\textwidth]{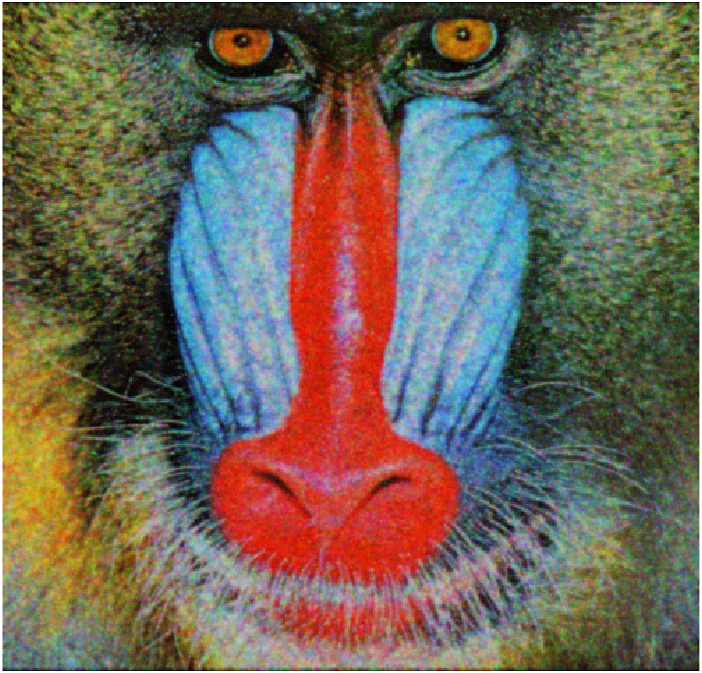}
        \caption{Model 1}
    \end{subfigure}
    \begin{subfigure}[b]{0.24\textwidth}
        \centering
        \includegraphics[width=\textwidth]{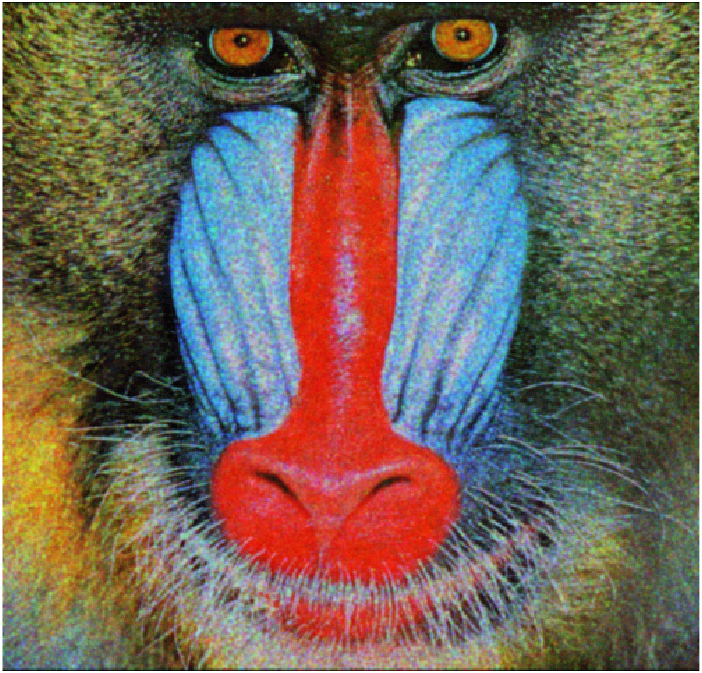}
        \caption{Model 1}
    \end{subfigure}

    \caption{
     The first Row contains noisy baboon images with noise level Look = 1, 3, 5, 10. 
        Subsequent rowss: Restored baboon images using different models (TDM, TDFM, Model 2, Model 1).
    }
    \label{fig:gamma_noise_results_baboon}
\end{figure}

Figure ~\ref{fig:gamma_noise_results_parrots} focuses on the Parrots grayscale image across different noise levels, arranged in the first row. The subsequent rows visualize the restoration results from four separate denoising models: the existing classical TDM Model\cite{29}, TDFM Model\cite{74}, and the two proposed models referred to as Model 1~\eqref{proposed_1} and Model 2~\eqref{proposed_2}, which employ weighted and coupled PDE approaches. Similarly, Figure~\ref {fig:gamma_noise_results_texture} replicates this layout for the ``Texture'' grayscale image. Figures ~\ref{fig:gamma_noise_results_caps} and ~\ref{fig:gamma_noise_results_baboon} present results for the color images Caps and Baboon, respectively. These visual comparisons, while insightful, are limited in providing a quantitative measure of denoising effectiveness because visual perceptions can be subjective and may not readily reveal subtle differences in performance.
To objectively evaluate and compare the denoising quality of these models, quantitative metrics such as the Mean Structural Similarity Index Measure (MSSIM) and the Peak Signal-to-Noise Ratio (PSNR) were calculated for each image and noise level. These results are tabulated in Table ~\ref{tab:comparison_gray} for grayscale images and Table ~\ref{tab:comparison_color} for color images. Notably, for the ``Parrots'' grayscale image, Model 1 yields the highest MSSIM scores, indicating superior structural preservation, while Model 2 achieves the best PSNR results, signifying more accurate intensity reconstruction. In contrast, for the ``Texture'' image, Model 1 demonstrates better texture preservation. Examining the trends across different noise levels, Model 1 prevails in MSSIM and PSNR for lower looks (1 and 3), suggesting it maintains both structure and intensity well under heavier noise conditions. Meanwhile, at higher looks (5 and 10), Model 1 excels in PSNR, but Model 2 slightly outperforms in MSSIM, emphasizing their strengths.

Table ~\ref{tab:comparison_color} shows comparative performance for various noise levels, revealing that Model 1 dominates MSSIM results at lower noise (looks 1 and 3), whereas Model 2 leads in PSNR for the same looks. For higher noise levels, Model 1 generally performs best across evaluated metrics. In the case of the ``Baboon'' image, characterized by complex textures and colors, Model 1 consistently achieves the best scores in both MSSIM and PSNR across all noise levels, reflecting its ability to retain intricate image details under various noise intensities.

Moreover, Tables ~\ref{tab:parameter_gray} and ~\ref{tab:parameter_color} provide detailed accounts of the optimal parameter settings ($\alpha, \beta, \gamma, \lambda$) employed in each model for the experiments. These parameters differ notably across image types and noise conditions, underscoring the necessity of adaptive tuning to obtain the best restoration results. The variation in parameter selection between grayscale and color images, as well as between different noise intensities, highlights the flexibility and robustness of the proposed PDE-based frameworks.

Table ~\ref{tab:si_sar_images_table} presents the optimal parameters for Models 1 and 2 that yield the best Speckle Index (S.I.) values, alongside the S.I. for both the noisy inputs and restored images. Two types of images were analyzed: Image 1, a Synthetic Aperture Radar (SAR) image, and Image 2, a medical ultrasound image. The results show that both models significantly reduce speckle noise, with Model 1 often achieving slightly better S.I. reductions.

\begin{figure}[H]
    \centering
    \begin{subfigure}[b]{0.25\textwidth}
        \centering
        \includegraphics[width=\textwidth, keepaspectratio]{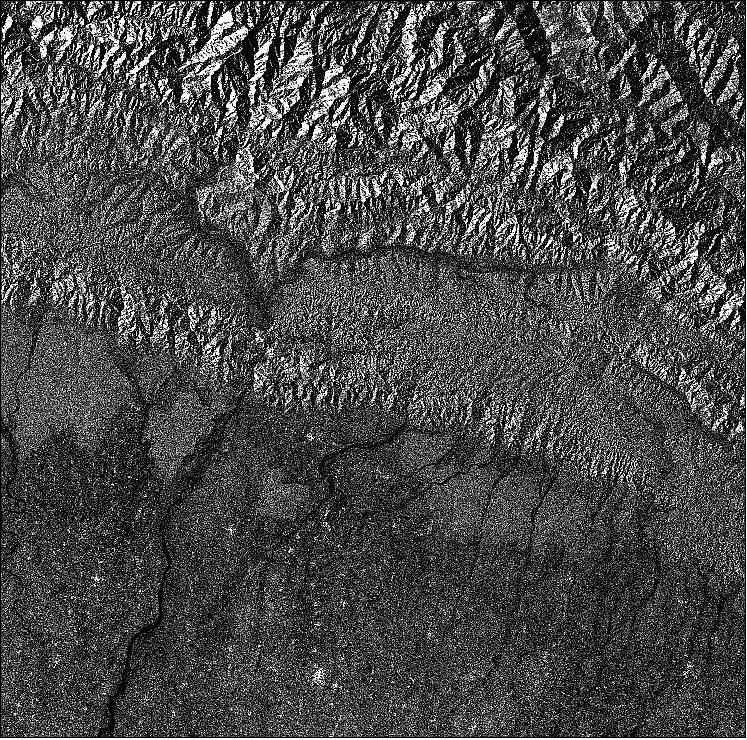}
        \caption*{ SAR Image}
    \end{subfigure}
    \begin{subfigure}[b]{0.25\textwidth}
        \centering
        \includegraphics[width=\textwidth, keepaspectratio]{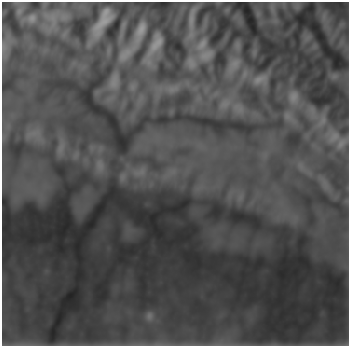}
        \caption*{Model 1 }
    \end{subfigure}
    \begin{subfigure}[b]{0.25\textwidth}
        \centering
        \includegraphics[width=\textwidth, keepaspectratio]{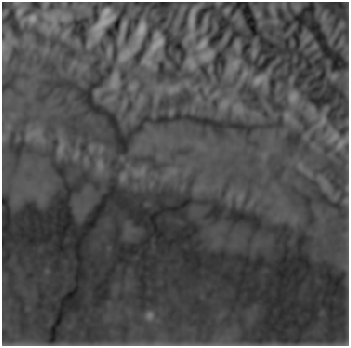}
        \caption*{Model 2 }
    \end{subfigure}

    \begin{subfigure}[b]{0.25\textwidth}
        \centering
        \includegraphics[width=\textwidth, keepaspectratio]{ULTRASOUND3.png}
        \caption*{Ultrasound Image}
    \end{subfigure}
    \begin{subfigure}[b]{0.25\textwidth}
        \centering
        \includegraphics[width=\textwidth, keepaspectratio]{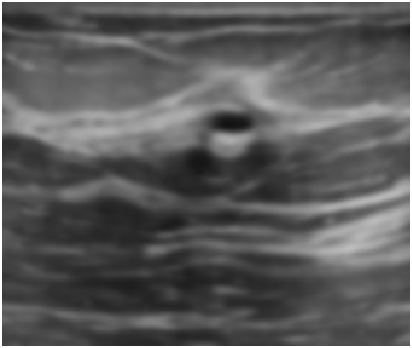}
        \caption*{Model 1}
    \end{subfigure}
    \begin{subfigure}[b]{0.25\textwidth}
        \centering
        \includegraphics[width=\textwidth, keepaspectratio]{coupled_ULTRASOUND3_FORTH_second_itratively_restored.png}
        \caption*{Model 2}
    \end{subfigure}

    \caption{Comparison of SAR Image Restoration. Each row presents    (1) Noisy Image (2) Model 1 Restored (3) Model 2 Restored.}
    \label{fig:sar_results}
\end{figure}

Figure ~\ref{fig:sar_results} visually demonstrates these improvements, displaying the noisy images alongside their restorations by Models 1 and 2 for both SAR and ultrasound. The restored images clearly exhibit diminished speckle patterns while preserving important structural details, confirming the quantitative S.I. improvements. These findings highlight the effectiveness and adaptability of the PDE-based denoising models across different imaging modalities, making them valuable for practical applications requiring noise suppression without loss of critical image information.

\begin{figure}[H]
    \centering

    \begin{subfigure}[b]{0.45\textwidth}
        \includegraphics[width=\linewidth]{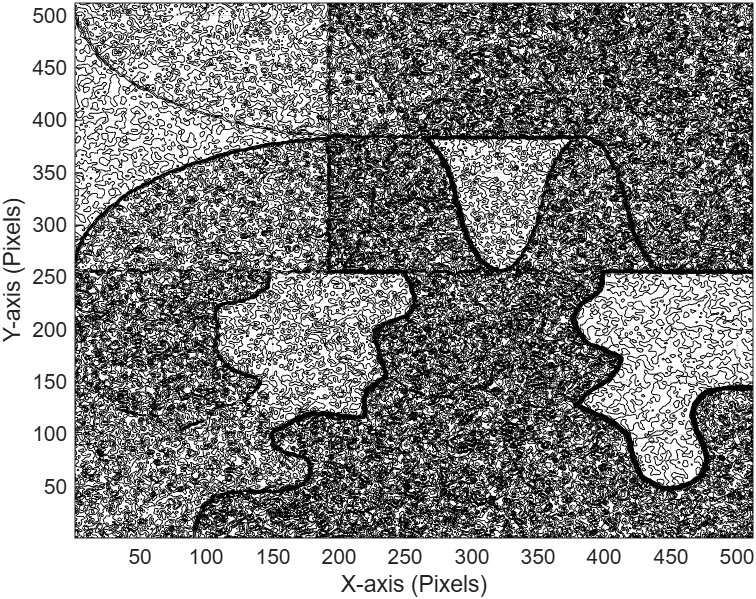}
        \caption{Noisy Image}
        \label{fig:noisy}
    \end{subfigure}
    \hspace{1em}
    \begin{subfigure}[b]{0.45\textwidth}
        \includegraphics[width=\linewidth]{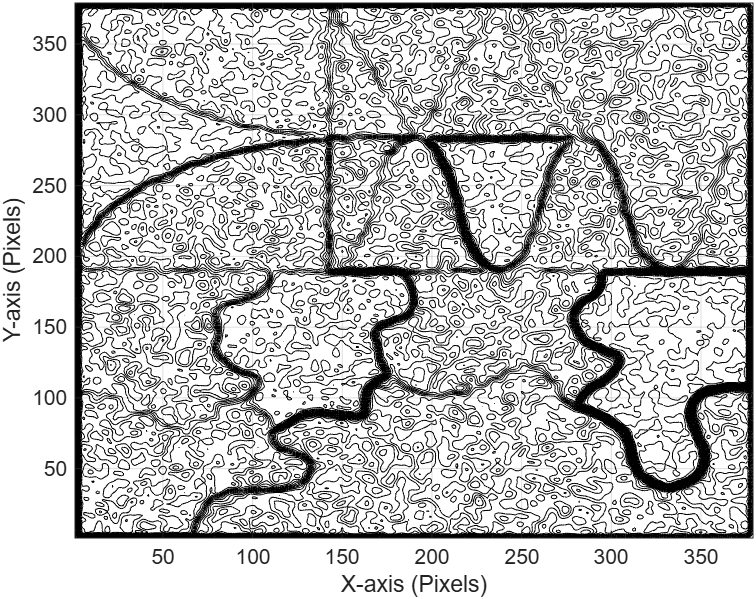}
        \caption{TDM Restored}
        \label{fig:shan}
    \end{subfigure}
    
    \vspace{1em}
    \begin{subfigure}[b]{0.45\textwidth}
        \includegraphics[width=\linewidth]{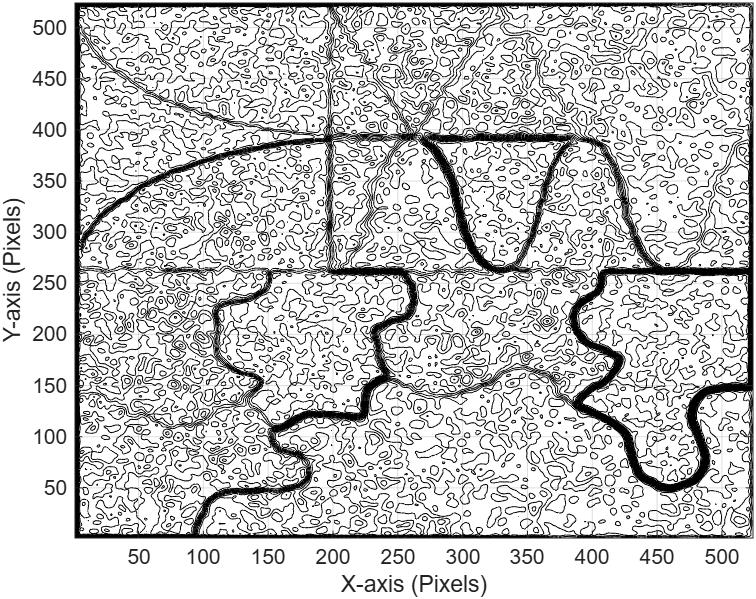}
        \caption{TDFM Restored}
        \label{fig:tde}
    \end{subfigure}
    \hspace{1em}
    \begin{subfigure}[b]{0.45\textwidth}
        \includegraphics[width=\linewidth]{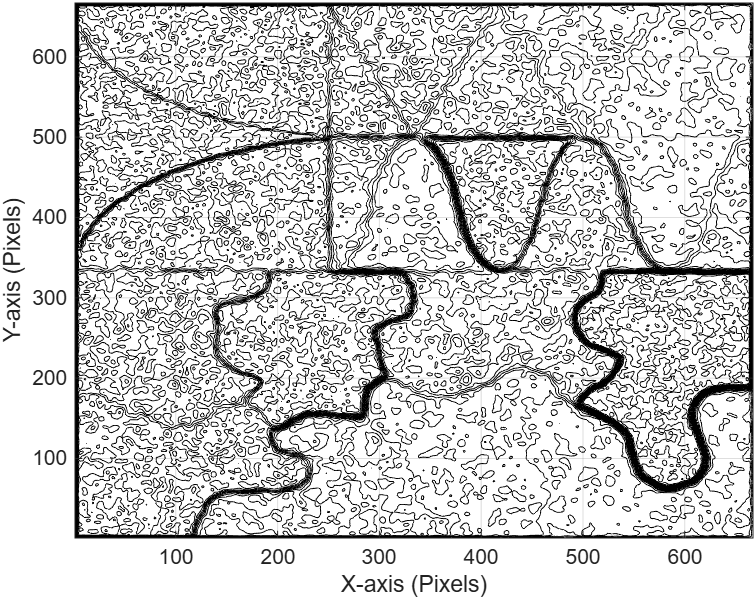}
        \caption{Model 2 Restored}
        \label{fig:proposed}
    \end{subfigure}

    \vspace{1em}
    \begin{subfigure}[b]{0.45\textwidth}
        \includegraphics[width=\linewidth]{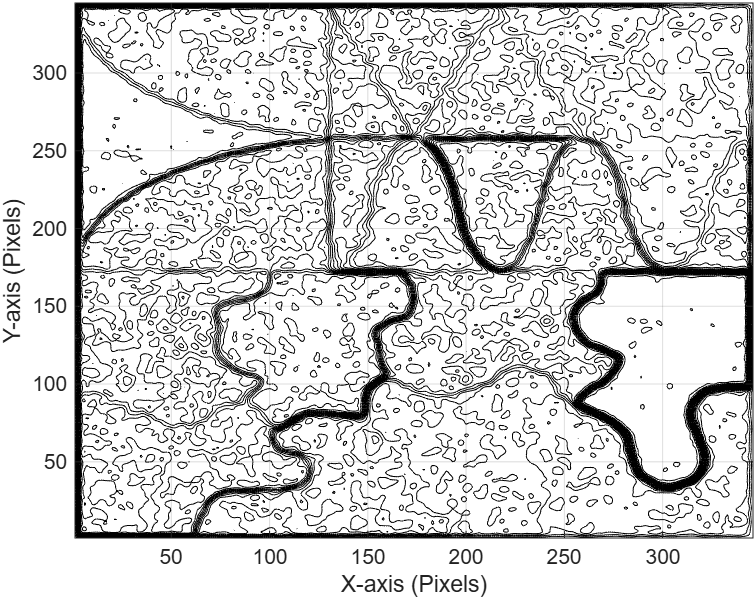}
        \caption{Model 1 Restored}
        \label{fig:gt}
    \end{subfigure}

    \caption{2D plots of texture gray image (Look=3) showing noisy, restored, and ground truth results.}
    \label{fig:2d_comparisons}
\end{figure}

\begin{figure}[H]
    \centering
   
    \begin{minipage}{0.45\textwidth}
        \centering
        \includegraphics[width=\linewidth]{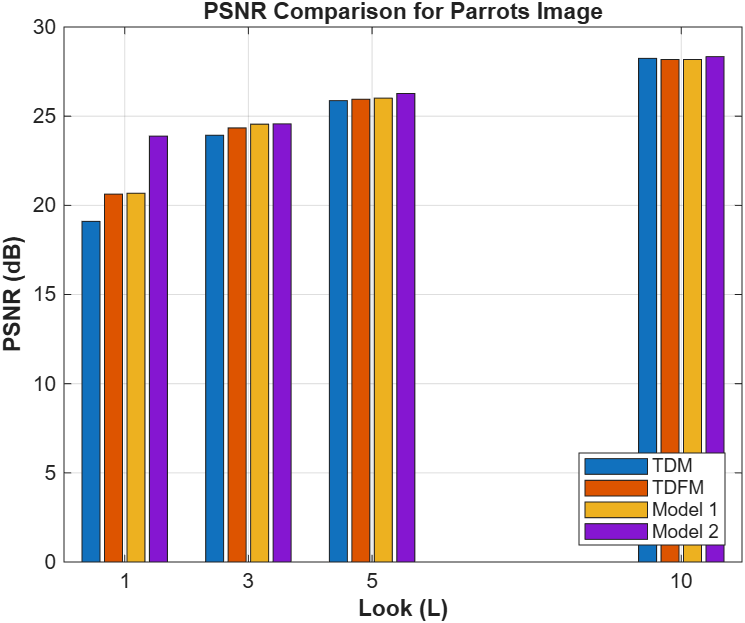}
        
    \end{minipage}
    \hfill
    \begin{minipage}{0.45\textwidth}
        \centering
        \includegraphics[width=\linewidth]{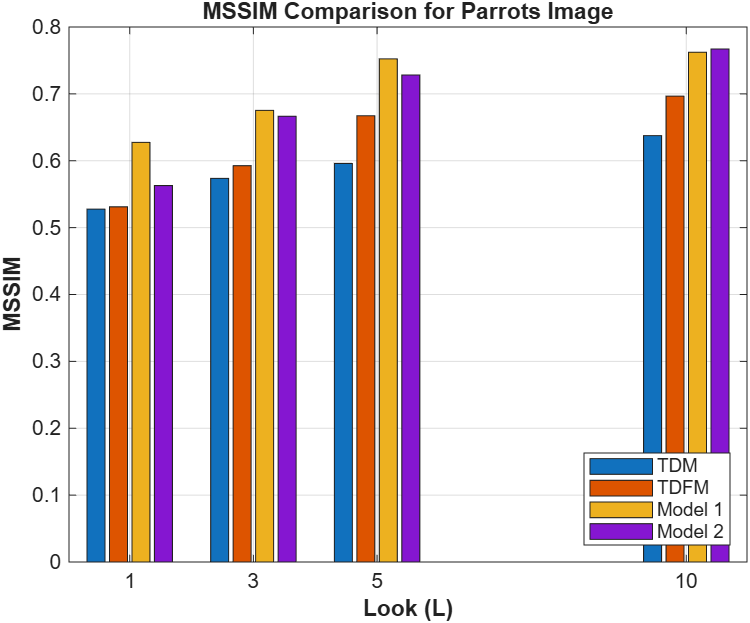}
       
    \end{minipage}
    
    \caption{Comparison of PSNR and MSSIM  Proposed Model 1, 2 and State-of-art TDM and TDFM for the gray parrots image}
     \label{fig:gray_PSNR}
\end{figure}

\begin{figure}[H]
    \centering
   
    \begin{minipage}{0.45\textwidth}
        \centering
        \includegraphics[width=\linewidth]{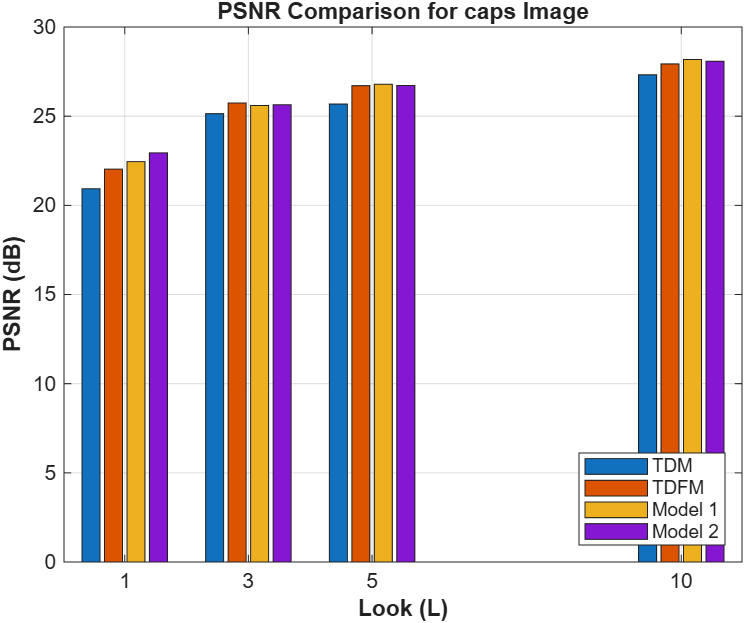}
        
    \end{minipage}
    \hfill
    \begin{minipage}{0.45\textwidth}
        \centering
        \includegraphics[width=\linewidth]{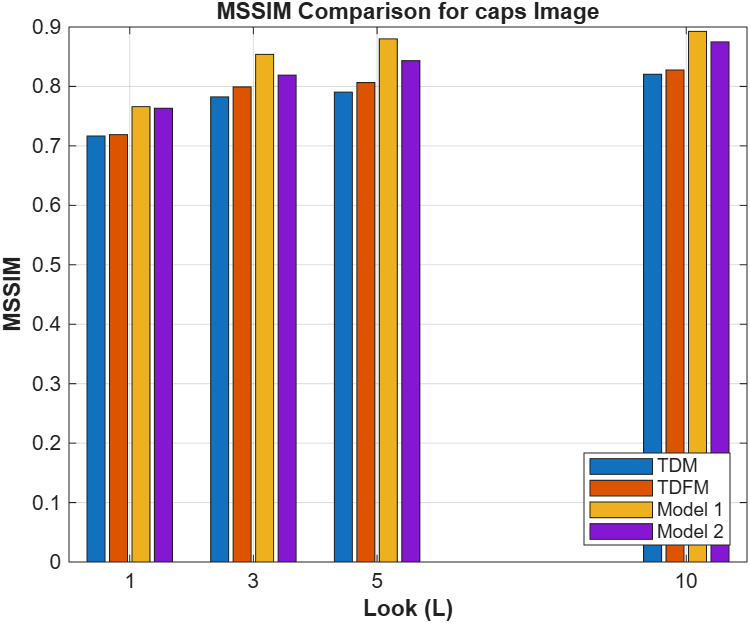}
       
    \end{minipage}
    
    \caption{Caparison of PSNR and MSSIM  Proposed Model 1, 2 and State-of-art TDM and TDFM for the Color caps image}
     \label{fig:color_PSNR}
\end{figure}

\begin{figure}[H]
    \centering
    
    \begin{subfigure}[b]{0.24\textwidth}
        \centering
        \includegraphics[width=\linewidth]{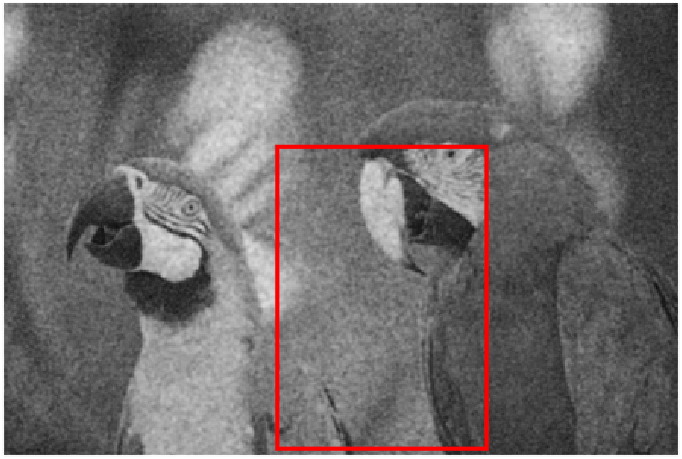}
        \caption{TDFM}
    \end{subfigure}
    \begin{subfigure}[b]{0.24\textwidth}
        \centering
        \includegraphics[width=\linewidth]{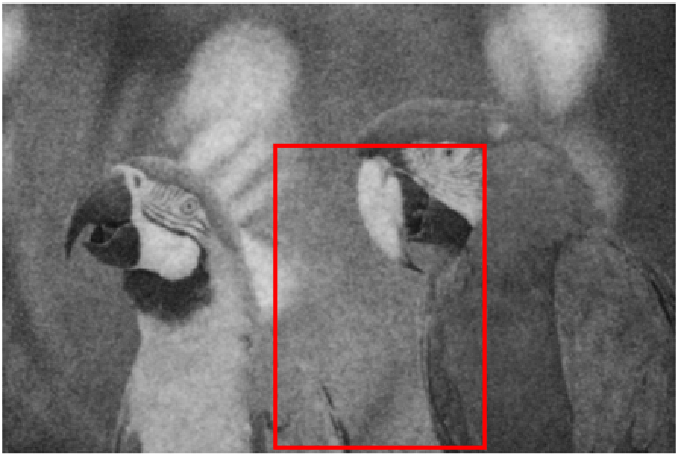}
        \caption{Model 1}
    \end{subfigure}
    \begin{subfigure}[b]{0.24\textwidth}
        \centering
        \includegraphics[width=\linewidth]{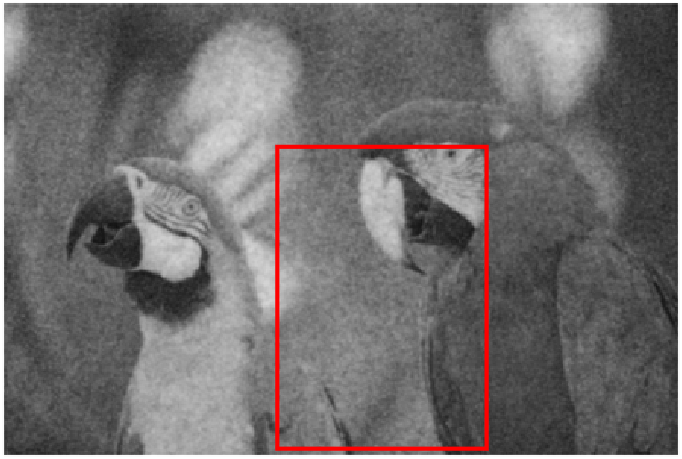}
        \caption{Model 2}
    \end{subfigure}

    \vspace{0.5cm} 

    \begin{subfigure}[b]{0.24\textwidth}
        \centering
        \includegraphics[width=\linewidth]{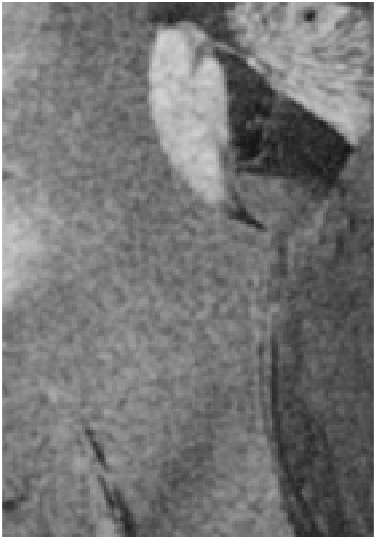}
        \caption{TDFM}
    \end{subfigure}
    \begin{subfigure}[b]{0.24\textwidth}
        \centering
        \includegraphics[width=\linewidth]{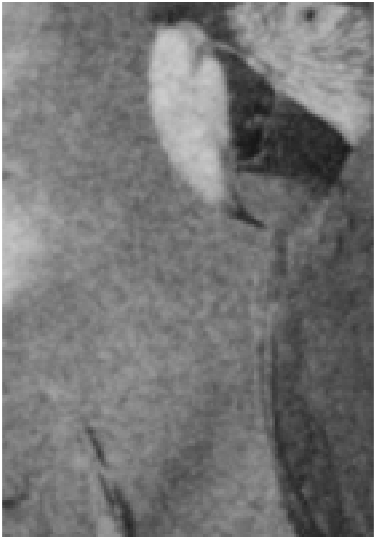}
        \caption{Model 1}
    \end{subfigure}
    \begin{subfigure}[b]{0.24\textwidth}
        \centering
        \includegraphics[width=\linewidth]{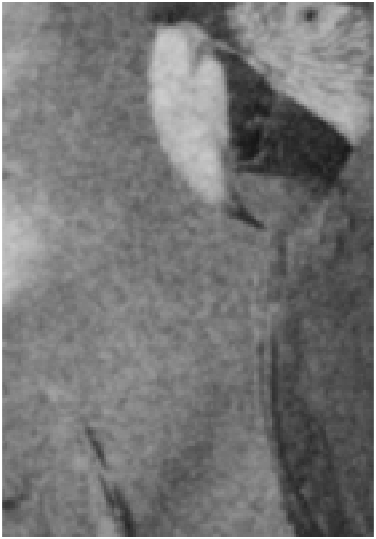}
        \caption{Model 2}
    \end{subfigure}

    \caption{Comparison of Denoising Performance: The first row shows the full parrots restored images(Look=3) with red bounding boxes highlighting the most de-speckled region. The second row presents the zoomed-in region of interest (ROI) from each image.}
    \label{fig:denoising_comparison_box}
    \end{figure}

Figure ~\ref{fig:2d_comparisons} displays the 2D contour plot for the "Texture" image at look 3, highlighting that Model 1 exhibits significantly less fluctuation in the contour representation. This reduced variability indicates that Model 1 achieves superior noise reduction compared to the other models for this noise level. Figures ~\ref{fig:gray_PSNR} and ~\ref{fig:color_PSNR} present the PSNR and SSIM metric trends across all look values for the "Parrots" and "Caps" images, respectively. These figures demonstrate consistent improvement by Models 1 and 2 over the classical Fourth Order telegraph model, illustrating the effectiveness of the proposed methods in preserving image structure and reducing noise quantitatively. Furthermore, Figure~\ref{fig:denoising_comparison_box} presents zoomed-in regions of interest (ROI) for the "Parrots" image, marked with bounding boxes, offering a detailed visual comparison of edge preservation and noise suppression. Within these ROIs, both Model 1 and Model 2 clearly outperform the TDFM Model by capturing sharper edges and reducing speckle more effectively, confirming their superiority in intricate image details restoration. Collectively, these visual and quantitative analyses underscore the advantage of the coupled PDE models, especially Model 1, in robustly addressing noise while retaining significant image features across different datasets and noise levels.

\begin{table}[H]
\centering
\caption{Comparison of PSNR and MSSIM values for different models on grayscale images.}
\label{tab:comparison_gray}
\begin{tabular}{|l|l|ll|ll|ll|ll|}
\hline
\multicolumn{1}{|c|}{Images} & \multicolumn{1}{c|}{Look} & \multicolumn{2}{c|}{TDM} & \multicolumn{2}{c|}{TDFM} & \multicolumn{2}{c|}{Model 1} & \multicolumn{2}{c|}{Model 2} \\ \hline
 &  & \multicolumn{1}{l|}{MSSIM} & PSNR & \multicolumn{1}{l|}{MSSIM} & PSNR & \multicolumn{1}{l|}{MSSIM} & PSNR & \multicolumn{1}{l|}{MSSIM} & PSNR \\ \hline
Parrots & \begin{tabular}[c]{@{}l@{}}1\\3\\5\\10\end{tabular} &
\multicolumn{1}{l|}{\begin{tabular}[c]{@{}l@{}}0.5277\\0.5737\\0.5960\\0.6376\end{tabular}} & 
\begin{tabular}[c]{@{}l@{}}19.10\\23.93\\25.87\\28.24\end{tabular} &
\multicolumn{1}{l|}{\begin{tabular}[c]{@{}l@{}}0.5311\\0.5926\\0.6673\\0.6966\end{tabular}} & 
\begin{tabular}[c]{@{}l@{}}20.63\\24.34\\25.95\\28.18\end{tabular} &
\multicolumn{1}{l|}{\begin{tabular}[c]{@{}l@{}}\textbf{0.6275}\\\textbf{0.6753}\\\textbf{0.7524}\\0.7623\end{tabular}} &
\begin{tabular}[c]{@{}l@{}}20.68\\24.55\\26.01\\28.18\end{tabular} &
\multicolumn{1}{l|}{\begin{tabular}[c]{@{}l@{}}0.5629\\0.6666\\0.7282\\\textbf{0.7671}\end{tabular}} &
\begin{tabular}[c]{@{}l@{}}\textbf{23.88}\\\textbf{24.57}\\\textbf{26.27}\\\textbf{28.34}\end{tabular} \\ \hline
Texture & \begin{tabular}[c]{@{}l@{}}1\\3\\5\\10\end{tabular} &
\multicolumn{1}{l|}{\begin{tabular}[c]{@{}l@{}}0.5284\\0.5534\\0.5727\\0.6010\end{tabular}} &
\begin{tabular}[c]{@{}l@{}}14.71\\19.30\\21.51\\24.11\end{tabular} &
\multicolumn{1}{l|}{\begin{tabular}[c]{@{}l@{}}0.5463\\0.6092\\0.6454\\0.7730\end{tabular}} &
\begin{tabular}[c]{@{}l@{}}17.20\\20.82\\21.48\\24.88\end{tabular} &
\multicolumn{1}{l|}{\begin{tabular}[c]{@{}l@{}}\textbf{0.6216}\\\textbf{0.7044}\\0.7203\\0.7846\end{tabular}} &
\begin{tabular}[c]{@{}l@{}}\textbf{17.29}\\20.84\\\textbf{22.58}\\\textbf{24.84}\end{tabular} &
\multicolumn{1}{l|}{\begin{tabular}[c]{@{}l@{}}0.5688\\0.6988\\\textbf{0.7525}\\\textbf{0.8270}\end{tabular}} &
\begin{tabular}[c]{@{}l@{}}17.28\\\textbf{20.86}\\21.54\\24.71\end{tabular} \\ \hline
\end{tabular}
\end{table}

\begin{table}[H]
\centering
\caption{Comparison of PSNR and MSSIM values for different models on color images.}
\label{tab:comparison_color}
\begin{tabular}{|l|l|ll|ll|ll|ll|}
\hline
\multicolumn{1}{|c|}{Images} & \multicolumn{1}{c|}{Look} & \multicolumn{2}{c|}{TDM} & \multicolumn{2}{c|}{TDFM} & \multicolumn{2}{c|}{Model 1} & \multicolumn{2}{c|}{Model 2} \\ \hline
 &  & \multicolumn{1}{l|}{MSSIM} & PSNR & \multicolumn{1}{l|}{MSSIM} & PSNR & \multicolumn{1}{l|}{MSSIM} & PSNR & \multicolumn{1}{l|}{MSSIM} & PSNR \\ \hline
caps & \begin{tabular}[c]{@{}l@{}}1\\3\\5\\10\end{tabular} &
\multicolumn{1}{l|}{\begin{tabular}[c]{@{}l@{}}0.7166\\0.7825\\0.7904\\0.8205\end{tabular}} &
\begin{tabular}[c]{@{}l@{}}20.93\\25.14\\25.68\\27.32\end{tabular} &
\multicolumn{1}{l|}{\begin{tabular}[c]{@{}l@{}}0.7190\\0.7992\\0.8065\\0.8277\end{tabular}} &
\begin{tabular}[c]{@{}l@{}}22.03\\25.74\\26.71\\27.93\end{tabular} &
\multicolumn{1}{l|}{\begin{tabular}[c]{@{}l@{}}\textbf{0.7660}\\\textbf{0.8540}\\\textbf{0.8800}\\\textbf{0.8926}\end{tabular}} &
\begin{tabular}[c]{@{}l@{}}22.45\\25.60\\\textbf{26.79}\\\textbf{28.18}\end{tabular} &
\multicolumn{1}{l|}{\begin{tabular}[c]{@{}l@{}}0.7633\\0.8190\\0.8433\\0.8750\end{tabular}} &
\begin{tabular}[c]{@{}l@{}}\textbf{22.94}\\\textbf{25.64}\\26.72\\28.08\end{tabular} \\ \hline
baboon & \begin{tabular}[c]{@{}l@{}}1\\3\\5\\10\end{tabular} &
\multicolumn{1}{l|}{\begin{tabular}[c]{@{}l@{}}0.5287\\0.6221\\0.6572\\0.7438\end{tabular}} &
\begin{tabular}[c]{@{}l@{}}16.79\\19.07\\19.80\\21.11\end{tabular} &
\multicolumn{1}{l|}{\begin{tabular}[c]{@{}l@{}}0.5687\\0.6821\\0.7366\\0.7998\end{tabular}} &
\begin{tabular}[c]{@{}l@{}}17.91\\19.51\\20.32\\21.60\end{tabular} &
\multicolumn{1}{l|}{\begin{tabular}[c]{@{}l@{}}\textbf{0.5925}\\\textbf{0.7197}\\\textbf{0.7626}\\\textbf{0.8187}\end{tabular}} &
\begin{tabular}[c]{@{}l@{}}\textbf{18.09}\\\textbf{20.13}\\\textbf{21.11}\\\textbf{22.29}\end{tabular} &
\multicolumn{1}{l|}{\begin{tabular}[c]{@{}l@{}}0.5696\\0.7089\\0.7613\\0.8151\end{tabular}} &
\begin{tabular}[c]{@{}l@{}}18.06\\19.93\\20.96\\22.10\end{tabular} \\ \hline
\end{tabular}
\end{table}

\begin{table}[H]
\centering
\caption{Parameter values for different models on grayscale images}
\label{tab:parameter_gray}
\begin{tabular}{|l|l|ccc|cccc|cccc|cccc|}
\hline
Image  & Look  & \multicolumn{3}{c|}{TDM} & \multicolumn{4}{c|}{TDFM} & \multicolumn{4}{c|}{Model 1} & \multicolumn{4}{c|}{Model 2} \\ \hline
       & L     & $\alpha$ & $\beta$ & $\gamma$ & $\alpha$ & $\beta$ & $\gamma$ & $\lambda$ & $\alpha$ & $\beta$ & $\gamma$ & $\lambda$ & $\alpha$ & $\beta$ & $\gamma$ & $\lambda$ \\ \hline

parrots   & \begin{tabular}[c]{@{}l@{}}1\\ 3\\ 5\\ 10\end{tabular} 
       & \begin{tabular}[c]{@{}l@{}}1\\1\\1\\1\end{tabular}
       & \begin{tabular}[c]{@{}l@{}}2\\2\\2\\2\end{tabular}
       & \begin{tabular}[c]{@{}l@{}}5\\5\\5\\5\end{tabular}

       & \begin{tabular}[c]{@{}l@{}}1\\ 1\\ 1\\ 1\end{tabular}    
       & \begin{tabular}[c]{@{}l@{}}3\\ 3\\ 3\\ 3\end{tabular}                
       & \begin{tabular}[c]{@{}l@{}}5\\ 5\\ 5\\ 5\end{tabular} 
       & \begin{tabular}[c]{@{}l@{}}0.05 \\0.1 \\0.1 \\0.2 \end{tabular} 

       & \begin{tabular}[c]{@{}l@{}}1\\ 1\\ 2\\ 1\end{tabular}       
       & \begin{tabular}[c]{@{}l@{}}2\\ 2\\ 2\\ 2\end{tabular}
       & \begin{tabular}[c]{@{}l@{}}2\\ 4\\ 4\\ 4\end{tabular} 
       & \begin{tabular}[c]{@{}l@{}}0.04\\ 0.07\\ 0.045\\ 0.1\end{tabular} 

       & \begin{tabular}[c]{@{}l@{}}1\\ 1\\ 2\\ 3\end{tabular}         
       & \begin{tabular}[c]{@{}l@{}}10\\ 10\\ 10\\ 10\end{tabular} 
       & \begin{tabular}[c]{@{}l@{}}1\\ 1\\ 1\\ 8\end{tabular}    
       & \begin{tabular}[c]{@{}l@{}}0.1\\ 0.1\\ 0.5\\ 0.1\end{tabular} \\ \hline

Texture  & \begin{tabular}[c]{@{}l@{}}1\\ 3\\ 5\\ 10\end{tabular} 
       & \begin{tabular}[c]{@{}l@{}}1\\1\\1\\1\end{tabular}
       & \begin{tabular}[c]{@{}l@{}}2\\2\\2\\2\end{tabular}
       & \begin{tabular}[c]{@{}l@{}}5\\5\\5\\5\end{tabular}

       & \begin{tabular}[c]{@{}l@{}}0.4\\ 1\\ 0.5\\ 0.5\end{tabular}            
       & \begin{tabular}[c]{@{}l@{}}3\\ 3\\ 3\\ 3\end{tabular}                
       & \begin{tabular}[c]{@{}l@{}}4\\ 4\\ 4\\ 4\end{tabular} 
       & \begin{tabular}[c]{@{}l@{}}0.03 \\0.03 \\0.03 \\0.05\end{tabular} 

       & \begin{tabular}[c]{@{}l@{}}1\\ 2\\ 0.2\\ 4\end{tabular}     
       & \begin{tabular}[c]{@{}l@{}}2\\ 2\\ 2\\ 2\end{tabular} 
       & \begin{tabular}[c]{@{}l@{}}2\\ 2\\ 1\\ 2\end{tabular} 
       & \begin{tabular}[c]{@{}l@{}}0.03\\ 0.03\\ 0.06\\ 0.035\end{tabular} 

       & \begin{tabular}[c]{@{}l@{}}2\\ 2\\ 2\\ 0.1\end{tabular}         
       & \begin{tabular}[c]{@{}l@{}}10\\ 10\\ 10\\ 10\end{tabular} 
       & \begin{tabular}[c]{@{}l@{}}1\\ 1\\ 2\\ 2\end{tabular}    
       & \begin{tabular}[c]{@{}l@{}}0.05\\ 0.05\\ 0.05\\ 0.05\end{tabular} \\ \hline

\end{tabular}
\end{table}

\begin{table}[H]
\centering
\caption{Parameter values for different models on color images}
\label{tab:parameter_color}
\begin{tabular}{|l|l|ccc|cccc|cccc|cccc|}
\hline
Image  & Look  & \multicolumn{3}{c|}{TDM} & \multicolumn{4}{c|}{TDFM} & \multicolumn{4}{c|}{Model 1} & \multicolumn{4}{c|}{Model 2} \\ \hline
       & L     & $\alpha$ & $\beta$ & $\gamma$ & $\alpha$ & $\beta$ & $\gamma$ & $\lambda$ & $\alpha$ & $\beta$ & $\gamma$ & $\lambda$ & $\alpha$ & $\beta$ & $\gamma$ & $\lambda$ \\ \hline

caps   & \begin{tabular}[c]{@{}l@{}}1\\3\\5\\10\end{tabular} 
       & \begin{tabular}[c]{@{}l@{}}3\\1\\2\\1.5\end{tabular} 
       & \begin{tabular}[c]{@{}l@{}}2\\2\\2\\2\end{tabular}
       & \begin{tabular}[c]{@{}l@{}}8\\8\\5\\5\end{tabular}

       & \begin{tabular}[c]{@{}l@{}}1\\1\\1\\1\end{tabular}    
       & \begin{tabular}[c]{@{}l@{}}3\\3\\3\\3\end{tabular}                
       & \begin{tabular}[c]{@{}l@{}}5\\5\\5\\5\end{tabular} 
       & \begin{tabular}[c]{@{}l@{}}0.5\\0.5\\0.5\\0.5\end{tabular}

       & \begin{tabular}[c]{@{}l@{}}1\\1\\1\\1\end{tabular}       
       & \begin{tabular}[c]{@{}l@{}}2\\2\\2\\2\end{tabular}
       & \begin{tabular}[c]{@{}l@{}}2\\1\\1\\1\end{tabular} 
       & \begin{tabular}[c]{@{}l@{}}0.08\\0.01\\0.01\\0.01\end{tabular} 

       & \begin{tabular}[c]{@{}l@{}}1\\1\\2\\3\end{tabular}         
       & \begin{tabular}[c]{@{}l@{}}10\\10\\10\\10\end{tabular} 
       & \begin{tabular}[c]{@{}l@{}}8\\8\\8\\8\end{tabular}    
       & \begin{tabular}[c]{@{}l@{}}0.08\\0.1\\0.1\\0.1\end{tabular} \\ \hline

baboon & \begin{tabular}[c]{@{}l@{}}1\\3\\5\\10\end{tabular} 
       & \begin{tabular}[c]{@{}l@{}}2\\2.5\\1.1\\1.2\end{tabular} 
       & \begin{tabular}[c]{@{}l@{}}2\\2\\2\\2\end{tabular}
       & \begin{tabular}[c]{@{}l@{}}9\\5\\5\\5\end{tabular}

       & \begin{tabular}[c]{@{}l@{}}1\\1.1\\1.2\\1.2\end{tabular}            
       & \begin{tabular}[c]{@{}l@{}}3\\3\\3\\3\end{tabular}                
       & \begin{tabular}[c]{@{}l@{}}5\\5\\5\\5\end{tabular} 
       & \begin{tabular}[c]{@{}l@{}}0.1\\0.1\\0.1\\0.1\end{tabular}

       & \begin{tabular}[c]{@{}l@{}}1\\1\\1\\1\end{tabular}     
       & \begin{tabular}[c]{@{}l@{}}2\\2\\2\\2\end{tabular} 
       & \begin{tabular}[c]{@{}l@{}}2\\2\\1\\2\end{tabular} 
       & \begin{tabular}[c]{@{}l@{}}0.1\\0.2\\0.2\\0.2\end{tabular} 

       & \begin{tabular}[c]{@{}l@{}}1\\1\\1\\1\end{tabular}         
       & \begin{tabular}[c]{@{}l@{}}10\\10\\10\\10\end{tabular} 
       & \begin{tabular}[c]{@{}l@{}}5\\5\\5\\5\end{tabular}    
       & \begin{tabular}[c]{@{}l@{}}0.1\\0.5\\0.5\\0.7\end{tabular} \\ \hline

\end{tabular}
\end{table}

\begin{table}[H]
\centering
\caption{Speckle Index (S.I.) results for Synthetic Aperture Radar (SAR) image and ultrasound image.
Model parameters are shown separately for Model 1 and Model 2.}
\label{tab:si_sar_images_table}
\begin{tabularx}{\textwidth}{c c *{4}{c} *{4}{c} c c}
\toprule
\textbf{Image} & \textbf{Noisy} &
\multicolumn{4}{c}{\textbf{Model 1 Parameters}} & 
\multicolumn{4}{c}{\textbf{Model 2 Parameters}} & 
\multicolumn{2}{c}{\textbf{S.I.}} \\
\cmidrule(lr){3-6}\cmidrule(lr){7-10}\cmidrule(lr){11-12}
 & & $\gamma$ & $\alpha$ & $\lambda$ & $K$ & 
     $\gamma$ & $\alpha$ & $\lambda$ & $K$ & 
     Model 1 & Model 2 \\
\midrule
Image\_1 & 1.02   & 5 & 1    & 0.001 & 2  & 5 & 1 & 0.0001 & 10 & \textbf{0.2210} & 0.2467 \\
Image\_2 & 0.5166 & 2 & 0.10 & 0.001 & 2  & 5 & 0.10 & 0.0001  & 10 & \textbf{0.4561} & 0.4696 \\
\bottomrule
\end{tabularx}
\end{table}

\section{Conclusion}

This study introduces and evaluates two novel PDE-based image denoising models—a weighted fourth-second PDE model (Model 1) and a coupled second and fourth order PDE model (Model 2), which clearly outperform the classical TDM and TDFM models across a variety of imaging modalities such as grayscale, color, SAR, and ultrasound. By independently tuning diffusion coefficients and employing adaptive coupling, these models achieve superior speckle noise suppression while effectively preserving critical image features, including edges, textures, and fine structures. Quantitative metrics, including PSNR, MSSIM, and speckle index, consistently demonstrate that both models significantly enhance noise reduction and structural fidelity, with Model 1 often providing the best overall performance. Visual assessments through multiple figures corroborate these quantitative results, showing sharper edges, clearer details, and reduced noise artifacts even at challenging noise levels.
While the current models focus mostly on the spatial aspects of diffusion and denoising quality, future research should explore improvements in the numerical time accuracy of these PDE-based solvers. Additionally, investigating adaptive timestep control and higher-order time integration methods may enhance robustness against noise variability and better preserve transient image features during restoration.
Overall, this work significantly advances the state-of-the-art in PDE-based image denoising by combining innovative mathematical modeling with thorough empirical validation and suggests promising directions to refine both algorithmic efficiency and restoration fidelity through improved numerical time accuracy in future studies.

\vspace{11pt}
\noindent\textbf{Author Declaration} \\
The authors declare that there are no conflicts of interest.

\vspace{11pt}
\noindent\textbf{Data Availability} \\
The data that support the findings of this study are available from the corresponding
author upon reasonable request.


\begin{thebibliography}{00}
\bibitem{1}
L. J. Porcello, N. G. Massey, R. B. Innes, and J. M. Marks, 
``Speckle reduction in synthetic-aperture radars,'' 
\emph{J. Opt. Soc. Amer.}, vol. 66, no. 11, pp. 1305--1311, Nov. 1976.

\bibitem{2}
F. Argenti, A. Lapini, L. Alparone, and T. Bianchi, 
``A tutorial on speckle reduction in synthetic aperture radar images,'' 
\emph{IEEE Geosci. Remote Sens. Mag.}, vol. 1, no. 3, pp. 6--35, Sep. 2013.

\bibitem{5}
J. W. Goodman, 
``Some fundamental properties of speckle,'' 
\emph{J. Opt. Soc. Amer.}, vol. 66, no. 11, pp. 1145--1150, Nov. 1976.

\bibitem{6}
M. K. Tur, C. Chin, and J. W. Goodman, 
``When is speckle noise multiplicative?'' 
\emph{Appl. Opt.}, vol. 21, no. 7, pp. 1157--1159, 1982.



\bibitem{8}
V. S. Frost, J. A. Stiles, K. S. Shanmugan, and J. C. Holtzman, 
``A model for radar images and its application to adaptive digital filtering of multiplicative noise,'' 
\emph{IEEE Trans. Pattern Anal. Mach. Intell.}, vol. PAMI-4, no. 2, pp. 157--166, Mar. 1982.

\bibitem{16} 
S. Parrilli, M. Poderico, C. V. Angelino, and L. Verdoliva, “A nonlocal
SAR image denoising algorithm based on LLMMSE wavelet shrink-
age,” IEEE Trans. Geosci. Remote Sens., vol. 50, no. 2, pp. 606–616,
Feb. 2012.
\bibitem{17}
H. Zhong, Y. Li, and L. Jiao, “SAR image despeckling using Bayesian
nonlocal means ﬁlter with sigma preselection,” IEEE Geosci. Remote
Sens. Lett., vol. 8, no. 4, pp. 809–813, Jul. 2011.
\bibitem{18}
C.-A. Deledalle, L. Denis, and F. Tupin, “Iterative weighted maximum
likelihood denoising with probabilistic patch-based weights,” IEEE
Trans. Image Process., vol. 18, no. 12, pp. 2661–2672, Dec. 2009.

\bibitem{21}
A. Achim, P. Tsakalides, and A. Bezerianos, “SAR image denoising
via Bayesian wavelet shrinkage based on heavy-tailed modeling,” IEEE
Trans. Geosci. Remote Sens., vol. 41, no. 8, pp. 1773–1784, Aug. 2003.

\bibitem{25}
S. Solbo and T. Eltoft, “Homomorphic wavelet-based statistical despeck-
ling of SAR images,” IEEE Trans. Geosci. Remote Sens., vol. 42, no. 4,
pp. 711–721, Apr. 2004.

\bibitem{27}
S. K. Jain and R. K. Ray, “Non-linear diffusion models for despeckling
of images: Achievements and future challenges,” IETE Tech. Rev.,
vol. 37, no. 1, pp. 66–82, 2020.
\bibitem{28}
S. K. Jain, R. K. Ray, and A. Bhavsar, “A nonlinear coupled diffusion
system for image despeckling and application to ultrasound images,”
Circuits, Syst., Signal Process., vol. 38, no. 4, pp. 1654–1683, 2019.
\bibitem{29}
S. Majee, R. K. Ray, and A. K. Majee, “A gray level indicator-based reg-
ularized telegraph diffusion model: Application to image despeckling,”
SIAM J. Imag. Sci., vol. 13, no. 2, pp. 844–870, Jan. 2020.
\bibitem{31}
J. Weickert, Anisotropic Diffusion in Image Processing, vol. 1. Stuttgart,
Germany: Teubner, 1998.
\bibitem{32}
Y. Yu and S. T. Acton, “Speckle reducing anisotropic diffusion,” IEEE
Trans. Image Process., vol. 11, no. 11, pp. 1260–1270, Nov. 2002.
\bibitem{34}
G. Aubert and J.-F. Aujol, “A variational approach to removing mul-
tiplicative noise,” SIAM J. Appl. Math., vol. 68, no. 4, pp. 925–946,
2008.
\bibitem{35}
L. Denis, F. Tupin, J. Darbon, and M. Sigelle, “SAR image regulariza-
tion with fast approximate discrete minimization,” IEEE Trans. Image
Process., vol. 18, no. 7, pp. 1588–1600, Jul. 2009.
\bibitem{36}
W. Feng, H. Lei, and Y. Gao, “Speckle reduction via higher order
total variation approach,” IEEE Trans. Image Process., vol. 23, no. 4,
pp. 1831–1843, Apr. 2014.
\bibitem{37}
S. Majee, S. K. Jain, R. K. Ray, and A. K. Majee, “A fuzzy edge detector driven telegraph total variation model for image despeckling,” Inverse Problems Imag., vol. 16, no. 2, pp. 367 396, 2022. [Online]. Available: https://www.aimsciences.org/article/doi/10.3934/ipi.2021054
\bibitem{38}
Z. Jin and X. Yang, “Analysis of a new variational model for multiplica-
tive noise removal,” J. Math. Anal. Appl., vol. 362, no. 2, pp. 415–426,
2010.
\bibitem{39} 
L. Rudin, P.-L. Lions, and S. Osher, “Multiplicative denoising and
deblurring: Theory and algorithms,” in Geometric Level Set Methods in
Imaging, Vision, and Graphics. New York, NY, USA: Springer, 2003,
pp. 103–119.
\bibitem{40}
J. Shi and S. Osher, “A nonlinear inverse scale space method for a
convex multiplicative noise model,” SIAM J. Imag. Sci., vol. 1, no. 3,
pp. 294–321, 2008.
\bibitem{41}
G. Chierchia, D. Cozzolino, G. Poggi, and L. Verdoliva, “SAR image
despeckling through convolutional neural networks,” in Proc. IEEE Int.
Geosci. Remote Sens. Symp. (IGARSS), Jul. 2017, pp. 5438–5441.
\bibitem{42}
A. B. Molini, D. Valsesia, G. Fracastoro, and E. Magli, “Speckle2Void:
Deep self-supervised SAR despeckling with blind-spot convolutional
neural networks,” IEEE Trans. Geosci. Remote Sens., vol. 60, pp. 1–17,
2022.
\bibitem{43}
A. G. Mullissa, D. Marcos, D. Tuia, M. Herold, and J. Reiche,
“DeSpeckNet: Generalizing deep learning-based SAR image
despeckling,” IEEE Trans. Geosci. Remote Sens., vol. 60, 2020,
Art. no. 5200315.
\bibitem{44}
P. Wang, H. Zhang, and V. M. Patel, “SAR image despeckling using
a convolutional neural network,” IEEE Signal Process. Lett., vol. 24,
no. 12, pp. 1763–1767, Dec. 2017.
\bibitem{45}
Y. Zhao, J. G. Liu, B. Zhang, W. Hong, and Y. R. Wu, “Adaptive total
variation regularization based SAR image despeckling and despeckling
evaluation index,” IEEE Trans. Geosci. Remote Sens., vol. 53, no. 5,
pp. 2765–2774, May 2015.
\bibitem{46}
L. Denis, E. Dalsasso, and F. Tupin, “A review of deep-learning
techniques for SAR image restoration,” in Proc. IEEE Int. Geosci.
Remote Sens. Symp., Jul. 2021, pp. 411–414.
\bibitem{47}
H. Cantalloube and C. Nahum, “How to compute a multi-look SAR
image?” Eur. Space Agency Publications, vol. 450, pp. 635–640,
Mar. 2000.
\bibitem{50}
K. Dabov, A. Foi, V. Katkovnik, and K. Egiazarian, “Image denoising
by sparse 3-D transform-domain collaborative ﬁltering,” IEEE Trans.
Image Process., vol. 16, no. 8, pp. 2080–2095, Aug. 2007.

\bibitem{65}
Cuomo, S., De Rosa, M., Izzo, S., Piccialli, F.,
 Pragliola, M. (2023).Speckle noise removal via learned variational models. 

\bibitem{68} 
Y.-L. You, M. Kaveh, Fourth-order partial differential equations for noise removal, 
\textit{IEEE Trans. Image Process.}, 9(10) (2000), pp. 1723–1730.
\bibitem{69}
S. Majee, R. K. Ray, and A. K. Majee, ``A new non-linear hyperbolic-parabolic coupled PDE model for image despeckling,'' \emph{IEEE Trans. Image Process.}, vol. 31, pp. 1963--1976, Feb. 2022.
\bibitem{70} 
P. Perona, J. Malik, Scale-space and edge detection using anisotropic diffusion, 
\textit{IEEE Trans. Pattern Anal. Mach. Intell.}, 12(7) (1990), pp. 629–639.
\bibitem{71} 
Z. Zhou, Z. Guo, D. Zhang, B. Wu, A nonlinear diffusion equation-based model for ultrasound speckle noise removal, 
\textit{J. Nonlinear Sci.}, 28 (2018), pp. 443–470.

\bibitem{72} 
Z. Zhou, Z. Guo, G. Dong, J. Sun, D. Zhang, B. Wu, A doubly degenerate diffusion model based on the gray level indicator for multiplicative noise removal,  \textit{IEEE Trans. Image Process.}, 24(1) (2014), pp. 249–260.
\bibitem{73} 
X. Shan, J. Sun, Z. Guo, Multiplicative noise removal based on the smooth diffusion equation, 
\textit{J. Math. Imaging Vision}, 61(6) (2019), pp. 763–779.
\bibitem{74}
R. K. Ray and M. Kumar, ``New fourth-order grayscale indicator-based telegraph diffusion model for image despeckling,'' \emph{arXiv preprint arXiv:2509.26010}, Sep. 2025.

\end{thebibliography}
\end{document}